\documentclass[utf8]{frontiers_arxiv} % for Science, Engineering and Humanities and Social Sciences articles
\usepackage{url,hyperref,lineno,microtype,subcaption}
\usepackage{color}
\usepackage{float}
\usepackage{booktabs}

\def\keyFont{\fontsize{8}{11}\helveticabold }
\def\firstAuthorLast{Mendelsohn {et~al.}} %use et al only if is more than 1 author
\def\Authors{Julia Mendelsohn\,$^{1,*}$, Yulia Tsvetkov\,$^{2}$ and Dan Jurafsky\,$^{3}$}
\def\Address{$^{1}$School of Information, University of Michigan, Ann Arbor , MI , USA \\
$^{2}$Language Technologies Institute, Carnegie Mellon University, Pittsburgh, PA, USA \\
$^{3}$Department of Linguistics, Stanford University, Stanford, CA, USA}

\def\corrAuthor{Julia Mendelsohn}

\def\corrEmail{juliame@umich.edu}

\begin{document}
\onecolumn
\firstpage{1}

\title[Dehumanization]{A framework for the computational linguistic analysis of dehumanization} 

\author[\firstAuthorLast ]{\Authors} %This field will be automatically populated
\address{} %This field will be automatically populated
\correspondance{} %This field will be automatically populated
\extraAuth{}
\maketitle
\begin{abstract}

Dehumanization is a pernicious psychological process that often leads to extreme intergroup bias, hate speech, and violence aimed at targeted social groups. Despite these serious consequences and the wealth of available data, dehumanization has not yet been computationally studied on a large scale. Drawing upon social psychology research, we create a computational linguistic framework for analyzing dehumanizing language by identifying linguistic correlates of salient components of dehumanization. We then apply this framework to analyze discussions of LGBTQ people in the \textit{New York Times} from 1986 to 2015. Overall, we find increasingly humanizing descriptions of LGBTQ people over time. However, we find that the label \textit{homosexual} has emerged to be much more strongly associated with dehumanizing attitudes than other labels, such as \textit{gay}. Our proposed techniques highlight processes of linguistic variation and change in discourses surrounding marginalized groups. Furthermore, the ability to analyze dehumanizing language at a large scale has implications for automatically detecting and understanding media bias as well as abusive language online.

\tiny
 \keyFont{ \section{Keywords:} computational sociolinguistics, dehumanization, lexical variation, language change, media, New York Times, LGBTQ} %All article types: you may provide up to 8 keywords; at least 5 are mandatory.

\end{abstract}
{\color{red} Trigger Warning: this paper contains offensive material that some may find upsetting, especially in Table 4 and Table 7.}

\section{Introduction}
Despite the American public's increasing acceptance of LGBTQ people and recent legal successes, LGBTQ individuals remain targets of hate and violence \citep{dinakar2012common,silva2016analyzing,gallup2019}. At the core of this issue is dehumanization, ``the act of perceiving or treating people as less than human'' \citep{haslam2016recent}, a process that heavily contributes to extreme intergroup bias \citep{haslam2006dehumanization}. Language is central to studying this phenomenon; like other forms of bias \citep{wiebe2004learning,greene2009more,recasens2013linguistic,voigt2017language,breitfeller-etal-2019-finding}, dehumanizing attitudes are expressed through subtle linguistic manipulations, even in carefully-edited texts. It is crucial to understand the use of such linguistic signals in mainstream media, as the media's representation of marginalized social groups has far-reaching implications for social acceptance, policy, and safety.

While small-scale studies of dehumanization and media representation of marginalized communities provide valuable insights (e.g. \cite{esses2013uncertainty}), there exist no known large-scale analyses, likely due to difficulties in quantifying such a subjective and multidimensional psychological process. However, the ability to do large-scale analysis is crucial for understanding how dehumanizing attitudes have evolved over long periods of time. Furthermore, by being able to account for a greater amount of media discourse, large-scale techniques can provide a more complete view of the media environment to which the public is exposed.

Linguistics and computer science offer valuable methods and insights on which large-scale techniques might be developed for the study of dehumanization. By leveraging more information about the contexts in which marginalized groups are discussed, computational linguistic methods enable large-scale study of a complex psychological phenomenon and can even reveal linguistic variations and changes not easily identifiable through qualitative analysis alone.

We develop a computational linguistic framework for analyzing dehumanizing language, with a focus on lexical signals of dehumanization. Social psychologists have identified numerous components of dehumanization, such as \textit{negative evaluations of a target group}, \textit{denial of agency}, \textit{moral disgust}, and \textit{likening members of a target group to non-human entities such as vermin}. Drawing upon this rich body of literature, we first identify linguistic analogs for these components and propose several computational techniques to measure these linguistic correlates. We then apply this general framework to explore changing representations of LGBTQ groups in the \textit{New York Times} over thirty years and both qualitatively and quantitatively evaluate our techniques within this case study. We additionally use this lens of dehumanization to investigate differences in social meaning between the denotationally-similar labels \textit{gay} and \textit{homosexual}. We focus on a single case study in order to conduct an in-depth analysis of our methodology, but our framework can generalize to study representations of other social groups, which we briefly explore in our discussion.

This paper aims to bridge the gaps between computational modeling, sociolinguistics, and dehumanization research with implications for several disciplines. In addition to enabling large-scale studies of dehumanizing language and media representation of marginalized social groups, these techniques can be built into systems that seek to capture both conscious and unconscious biases in text. Furthermore, this work has implications for improving machines' abilities to automatically detect hate speech and abusive language online, which are typically underpinned by dehumanizing language. Finally, our case study demonstrates that such computational analyses of discussions about marginalized groups can provide unique insights into language variation and change within sensitive sociopolitical contexts, and help us understand how people (and institutions) use language to express their ideologies and attitudes towards certain social groups.

\section{Background}
\subsection{Dehumanization \label{sec:components}}

Our lexical semantic analysis involves quantifying linguistic correlates of component psychological processes that contribute to dehumanization. Our approaches are informed by social psychology research on dehumanization, which is briefly summarized here. Prior work has identified numerous related processes that comprise dehumanization \citep{haslam2006dehumanization}. One such component is \emph{likening members of the target group to non-human entities}, such as machines or animals \citep{haslam2006dehumanization,goff2008not,kteily2015ascent}. By perceiving members of a target group to be non-human, they are ``outside the boundary in which moral values, rules, and considerations of fairness apply" \citep{opotow1990moral}, which thus leads to violence and other forms of abuse. Metaphors and imagery relating target groups to vermin are particularly insidious and played a prominent role in the genocide of Jews in Nazi Germany and Tutsis in Rwanda \citep{harris2015dehumanized}. More recently, the vermin metaphor has been invoked by the media to discuss terrorists and political leaders of majority-Muslim countries after September 11 \citep{steuter2010vermin}. According to \cite{tipler2014agency}, the vermin metaphor is particularly powerful because it conceptualizes the target group as ``engaged in threatening behavior, but devoid of thought or emotional desire". 

\emph{Disgust} underlies the dehumanizing nature of these metaphors and is itself another important element of dehumanization. Disgust contributes to members of target groups being perceived as less-than-human and of negative social value \citep{sherman2011cuteness}. It is often evoked (both in real life and experimental settings) through likening a target group to animals. \cite{buckels2013disgust} find that priming participants to feel disgust facilitates ``moral exclusion of out-groups". Experiments by \cite{sherman2011cuteness} and \cite{hodson2007interpersonal} similarly find that disgust is a predictor of dehumanizing perceptions of a target group. Both moral disgust towards a particular social group and the invocation of non-human metaphors are facilitated by \textit{essentialist} beliefs about groups, which \cite{haslam2006dehumanization} presents as a necessary component of dehumanization. In order to distinguish between human and non-human, dehumanization requires an exaggerated perception of intergroup differences. Essentialist thinking thus contributes to dehumanization by leading to the perception of social groups as categorically distinct, which in turn emphasizes intergroup differences \citep{haslam2006dehumanization}

According to \cite{haslam2006dehumanization}, prior work describes \emph{``extremely negative evaluations of others"} as a major component of dehumanization. This is especially pronounced in \citeauthor{bar1990causes}'s account of delegitimization, which involves using negative characteristics to categorize groups that are ``excluded from the realm of acceptable norms and values" \citeyearpar{bar1990causes}. While \citeauthor{bar1990causes} defines delegitimization as a distinct process, he considers dehumanization to be one means of delegitimization. \cite{opotow1990moral} also discusses broader processes of moral exclusion, one of which is dehumanization. A closely related process is \textit{psychological distancing}, in which one perceives others to be objects or nonexistent \citep{opotow1990moral}. \cite{nussbaum1999sex} identifies elements that contribute to the objectification (and thus dehumanization) of women, one of which is \textit{denial of subjectivity}, or the habitual neglect of one's experiences, emotions, and feelings.

Another component of dehumanization is the \textit{denial of agency} to members of the target group \citep{haslam2006dehumanization}. According to \citeauthor{tipler2014agency}, there are three types of agency: "the ability to (1) experience emotion and feel pain (affective mental states), (2) act and produce an effect on their environment (behavioral potential), and (3) think and hold beliefs (cognitive mental states) \citeyearpar{tipler2014agency}. Dehumanization typically involves the denial of one or more of these types of agency 
\citep{tipler2014agency}.

In Section \ref{sec:operationalizing}, we introduce computational linguistic methods to quantify several of these components. 

\subsection{Related Computational Work}

While this is the first known computational work that focuses on dehumanization, we draw upon a growing body of literature at the intersection of natural language processing and social science. We are particularly inspired by the area of automatically detecting subjective language, largely pioneered by Janyce Wiebe and colleagues who developed novel lexical resources and algorithms for this task \citep{wiebe2004learning}. These resources have been used as linguistically-informed features in machine learning classification of biased language \citep{recasens2013linguistic}. Other work has expanded this lexicon-based approach to account for the role of syntactic form in identifying the writer's perspective towards different entities \citep{greene2009more, rashkin2016connotation}. 

These methods have been used and expanded to analyze pernicious, but often implicit social biases \citep{caliskan2017semantics}. For example, \citeauthor{voigt2017language} analyze racial bias in police transcripts by training classifiers with linguistic features informed by politeness theory \citeyearpar{voigt2017language}, and \citeauthor{garg2018word} investigate historical racial biases through changing word embeddings \citeyearpar{garg2018word}. Other studies focus on how people's positions in different syntactic contexts affect power and agency, and relate these concepts to gender bias in movies \citep{sap2017connotation} and news articles about the \#MeToo movement \citep{field2019contextual}. There is also a growing focus on identifying subtle manifestations of social biases, such as condescension \citep{wang2019talkdown}, microagressions \citep{breitfeller-etal-2019-finding}, and ``othering" language \citep{burnap2016us,alorainy2019enemy}. In addition, our focus on dehumanization is closely related to the detection and analysis of hate speech and abusive language \citep{schmidt2017survey,elsherief2018hate}.

Gender and racial bias have also been identified within widely-deployed NLP systems, for tasks including toxicity detection \citep{sap2019social}, sentiment analysis \citep{kiritchenko2018examining}, coreference resolution \citep{rudinger2018gender}, language identification \citep{blodgett2017racial}, and in many other areas \citep{sun2019mitigating}. Given the biases captured, reproduced, and perpetuated in NLP systems, there is a growing interest in mitigating subjective biases \citep{sun2019mitigating}, with approaches including modifying embedding spaces \citep{bolukbasi2016man,manzini2019black}, augmenting datasets \citep{zhao2018gender}, and adapting natural language generation methods to ``neutralize" text \citep{pryzant2019automatically}.

A related line of research has developed computational approaches to investigate language use and variation in media discourse about sociopolitical issues. For example, some work has drawn upon political communication theory to automatically detect an issue's framing \citep{entman1993framing, boydstun2013identifying,card2015media} through both supervised classification \citep{boydstun2014tracking,baumer2015testing} and unsupervised methods, such as topic modeling and lexicon induction \citep{tsur2015frame,field2018framing,demszky2019analyzing}. Scholars have also developed computational methods to identify lexical cues of partisan political speech, political slant in mass media, and polarization in social media \citep{monroe2008fightin,gentzkow2010drives,demszky2019analyzing}.

\subsection{Attitudes Towards LGBTQ Communities in the United States}

Some background about LGBTQ communities is necessary for our case study of LGBTQ dehumanization in the \textit{New York Times}. Bias against LGBTQ people is longstanding in the United States. Overall, however, the American public has become more accepting of LGBTQ people and supportive of their rights. In 1977, equal percentages of respondents (43\%) agreed and disagreed with the statement that gay or lesbian relations between consenting adults should be legal \citep{gallup2019}. Approval of gay and lesbian relations then decreased in the 1980s; in 1986, only 32\% of respondents believed they should be legal. According to Gallup, attitudes have become increasingly positive since the 1990s, and in 2019, 73\% responded that gay or lesbian relations should be legal. The Pew Research center began surveying Americans about their beliefs about same-sex marriage in 2001 and found similar trends \citep{PewResearchCenter}. Between 2001 and 2019, support for same-sex marriage jumped from 35\% to 61\%.

In addition to the public's overall attitudes, it is important to consider the specific words used to refer to LGBTQ people. Because different group labels potentially convey different social meanings, and thus have different relationships with dehumanization, our case study compares two LGBTQ labels: \textit{gay} and \textit{homosexual}. The Gallup survey asked for opinions on legality of ``homosexual relations" until 2008, but then changed the wording to ``gay and lesbian relations". This was likely because many gay and lesbian people find the word \textit{homosexual} to be outdated and derogatory. According to the LGBTQ media monitoring organization GLAAD, \textit{homosexual}'s offensiveness originates in the word's dehumanizing clinical history, which had falsely suggested that ``people attracted to the same sex are somehow diseased or psychologically/emotionally disordered" \footnote{https://www.glaad.org/reference/lgbtq}. Beyond its outdated clinical associations, some argue that the word \textit{homosexual} is more closely associated with sex and all of its negative connotations simply by virtue of containing the word \textit{sex}, while terms such as \textit{gay} and \textit{lesbian} avoid such connotations \citep{peters_2014}. Most newspapers, including the \textit{New York Times}, almost exclusively used the word \textit{homosexual} in articles about gay and lesbian people until the late 1980s \citep{soller_2018}. The \textit{New York Times} began using the word \textit{gay} in non-quoted text in 1987. Many major newspapers began restricting the use of the word \textit{homosexual} in 2006 \citep{peters_2014}. As of 2013, the \textit{New York Times} has confined the use of \textit{homosexual} to specific references to sexual activity or clinical orientation, in addition to direct quotes and paraphrases \footnote{https://www.glaad.org/reference/style}.

Beyond differences in how LGBTQ people perceive the terms \textit{gay} or \textit{lesbian} relative to \textit{homosexual}, the specific choice of label can affect attitudes towards LGBTQ people. In 2012, \cite{smith2017gay} asked survey respondents about either ``gay and lesbian rights" or ``homosexual rights". Respondents who read the word ``homosexual" showed less support for LGBTQ rights. This effect was primarily driven by high authoritarians, people who show high sensitivity to intergroup distinctions. The authors posit that \textit{homosexual} makes social group distinctions more blatant than \textit{gay} or \textit{lesbian}. This leads to greater psychological distancing, thus enabling participants to remove LGBTQ people from their realm of moral consideration \citep{smith2017gay}. Based on prior research and evolving media guidelines, we expect our computational analysis to show that \textit{homosexual} occurs in more dehumanizing contexts than the label \textit{gay}.

\section{Operationalizing Dehumanization} \label{sec:operationalizing}

In Section \ref{sec:components}, we discussed multiple elements of dehumanization that have been identified in social psychology literature. Here we introduce and quantify lexical correlates to operationalize four of these components: \textit{negative evaluations of a target group}, \textit{denial of agency}, \textit{moral disgust}, and \textit{use of vermin metaphors}. 

\subsection{Negative Evaluation of a Target Group \label{sec:sentiment}}
One prominent aspect of dehumanization is extremely negative evaluations of members of a target group \citep{haslam2006dehumanization}. Attribution of negative characteristics to members of a target group in order to exclude that group from ``the realm of acceptable norms and values" is specifically the key component of \textit{delegitimization}, a process of moral exclusion closely related to dehumanization. We hypothesize that this negative evaluation of a target group can be realized by words and phrases whose connotations have extremely low valence, where valence refers to the dimension of meaning corresponding to positive/negative (or pleasure/displeasure) \citep{osgood1957measurement,vad-acl2018}. Thus, we propose several valence lexicon-based approaches to measure this component: paragraph-level valence analysis, Connotation Frames of perspective, and word embedding neighbor valence. Each technique has different advantages and drawbacks regarding precision and interpretability.

\subsubsection{Paragraph-level Valence Analysis}

One dimension of affective meaning is \textit{valence}, which corresponds to an individual's evaluation of an event or concept, ranging from negative/unpleasant to positive/pleasant \citep{osgood1957measurement,russell1980circumplex}. A straightforward lexical approach to measure \textit{negative evaluations of a target group} involves calculating the average valence of words occurring in discussions of the target group. We obtain valence scores for 20,000 words from the NRC VAD lexicon, which contains real-valued scores ranging from zero to one for valence, arousal and dominance. A score of zero represents the lowest valence (most negative emotion) and a score of one is the highest possible valence (most positive emotion) \citep{vad-acl2018}. Words with the highest valence include \textit{love} and \textit{happy}, while words with the lowest valence include \textit{nightmare} and \textit{shit}.

We use paragraphs as the unit of analysis because a paragraph represents a single coherent idea or theme \citep{hinds1977paragraph}. This is particularly true for journalistic writing \citep{shuman1894steps}, and studies on rhetoric in journalism often treat paragraphs as the unit of analysis (e.g. \citealp{barnhurst1997american,katajamaki2006rhetorical}). Furthermore, by looking at a small sample of our data, we found that paragraphs were optimal because full articles often discuss unrelated topics while single sentences do not provide enough context to understand how the newspaper represents the target group. We calculate paragraph-level scores by taking the average valence score over all words in the paragraph that appear (or whose lemmas appear) in the NRC VAD lexicon.

\subsubsection{Connotation Frames of Perspective}

While paragraph-level valence analysis is straightforward, it is sometimes too coarse because we aim to understand the sentiment \textit{directed towards} the target group, not just nearby in the text. For example, suppose the target group is named ``B". A sentence such as ``A violently attacked B" would likely have extremely negative valence, but the writer may not feel negatively towards the victim, ``B".  

We address this by using \citeauthor{rashkin2016connotation}'s Connotation Frames Lexicon, which contains rich annotations for 900 English verbs (\citeyear{rashkin2016connotation}). Among other things, for each verb, the Connotation Frames Lexicon provides scores (ranging from -0.87 to 0.8) for the writer's perspective towards the verb's subject and object. In the example above for the verb \textit{attack}, the lexicon lists the writer's perspective towards the subject ``A", the attacker, as -0.6 (strongly negative) and the object ``B" as 0.23 (weakly positive). 

We extract all subject-verb-object tuples containing at least one target group label using the Spacy dependency parser \footnote{spacy.io}. For each subject and object, we capture the noun and the modifying adjectives, as group labels (such as \textit{gay}) can often take either nominal or adjectival forms. For each tuple, we use the Connotation Frames lexicon to determine the writer's perspective towards the noun phrase containing the group label. We then average perspective scores over all tuples.

\subsubsection{Word Embedding Neighbor Valence \label{sec:embed}}

While a Connotation Frames approach can be more precise than word-counting valence analysis, it limits us to analyzing SVO triples, which excludes a large portion of the available data about the target groups. This reveals a conundrum: broader context can provide valuable insights into the implicit evaluations of a social group, but we also want to directly probe attitudes towards the group itself. 

We address this tension by training vector space models to represent the data, in which each unique word in a large corpus is represented by a vector (embedding) in high-dimensional space. The geometry of the resulting vector space captures many semantic relations between words. Furthermore, prior work has shown that vector space models trained on corpora from different time periods can capture semantic change \citep{kulkarni2015statistically,hamilton_semchange}. For example, diachronic word embeddings reveal that the word \textit{gay} meant ``cheerful" or ``dapper" in the early $20^{\text{th}}$ century, but shifted to its current meaning of sexual orientation by the 1970s. Because word embeddings are created from real-world data, they contain real-world biases. For example, \cite{bolukbasi2016man} demonstrated that gender stereotypes are deeply ingrained in these systems. Though problematic for the widespread use of these models in computational systems, these revealed biases indicate that word embeddings can actually be used to identify stereotypes about social groups and understand how they change over time \citep{garg2018word}. 

This technique can similarly be applied to understand how a social group is negatively evaluated within a large text corpus. If the vector corresponding to a social group label is located in the semantic embedding space near words with clearly negative evaluations, that group is likely negatively evaluated (and possibly dehumanized) in the text.  

We first preprocess the data by lowercasing, removing numbers, and removing punctuation. We then use the word2vec skip-gram model to create word embeddings \citep{mikolov2013distributed}. We use Gensim's default parameters with two exceptions; we train our models for ten iterations in order to ensure that the models converge to the optimal weights and we set the window size to 10 words, as word vectors trained with larger window sizes tend to capture more semantic relationships between words \citep{levy2014dependency}\footnote{https://radimrehurek.com/gensim/models/word2vec.html}. For our diachronic analysis, we first train word2vec on the entire corpus, and then use the resulting vectors to initialize word2vec models for each year of data in order to encourage coherence and stability across years. After training word2vec, we zero-center and normalize all embeddings to alleviate the hubness problem \citep{dinu2014improving}. 

We then identify vectors for group labels by taking the centroid of all morphological forms of the label, weighted by frequency. For example, the vector representation for the label \textit{gay} is actually the weighted centroid of the words \textit{gay} and \textit{gays}. This enables us to simultaneously account for adjectival, singular nominal, and plural nominal forms for each social group label with a single vector. Finally, we estimate the valence for each group label by identifying its 500 nearest neighbors via cosine similarity, and calculating the average valence of all neighbors that appear in the NRC VAD Valence Lexicon \footnote{We conducted additional analyses by considering 25, 50, 100, 250, and 1000 nearest neighbors, which yielded similar results and can be found in the Supplementary Material.}.  

We also induce a valence score directly from a group label's vector representation by adapting the regression-based sentiment prediction from \citet{field2019entity} for word embeddings. This approach yielded similar results as analyzing nearest neighbor valence but was difficult to interpret. More details for and results from this technique can be found in the Supplementary Material.

\subsection{Denial of Agency}

\textit{Denial of agency} refers to the lack of attributing a target group member with the ability to control their own actions or decisions \citep{tipler2014agency}. Automatically detecting the extent to which a writer attributes cognitive abilities to a target group member is an extraordinarily challenging computational task. Fortunately, the same lexicons used to operationalize \textit{negative evaluations} provide resources for measuring lexical signals of \textit{denial of agency}. 

\subsubsection{Connotation Frames}
As in Section \ref{sec:sentiment}, we use Connotation Frames to quantify the amount of agency attributed to a target group. We use \citeauthor{sap2017connotation}'s extension of Connotation Frames for agency \citeyearpar{sap2017connotation}. Following \citeauthor{sap2017connotation}'s interpretation, entities with high agency exert a high degree of control over their own decisions and are active decision-makers, while entities with low agency are more passive \citeyearpar{sap2017connotation}. This contrast is particularly apparent in example sentences such as \textit{X searched for Y} and \textit{X waited for Y}, where the verb \textit{searched} gives X high agency and \textit{waited} gives X low agency \citep{sap2017connotation}. Additionally, \citeauthor{sap2017connotation}'s released lexicon for agency indicates that subjects of verbs such as \textit{attack} and \textit{praise} have high agency, while subjects of \textit{doubts} and \textit{needs} have low agency \citeyearpar{sap2017connotation}. 

This lexicon considers the agency attributed to subjects of nearly 2000 transitive and intransitive verbs. To use this lexicon to quantify \textit{denial of agency}, we extract all sentences' head verbs and their subjects, where the subject noun phrase contains a target group label. Unlike \citeauthor{rashkin2016connotation}'s real-valued Connotation Frames lexicon for perspective, the agency lexicon only provides binary labels, so we calculate the fraction of subject-verb pairs where the subject has high agency.

\subsubsection{Word Embedding Neighbor Dominance}

The NRC VAD Dominance Lexicon provides another resource for quantifying dehumanization \cite{vad-acl2018}. The NRC VAD lexicon's dominance dimension contains real-valued scores between zero and one for 20,000 English words. However, the dominance lexicon primarily captures power, which is distinct from but closely related to agency. While power refers to one's control over others, agency refers to one's control over oneself. While this lexicon is a proxy, it qualitatively appears to capture signals of \textit{denial of agency}; the highest dominance words are \textit{powerful}, \textit{leadership}, \textit{success}, and \textit{govern}, while the lowest dominance words are \textit{weak}, \textit{frail}, \textit{empty}, and \textit{penniless}. We thus take the same approach as in Section \ref{sec:embed}, but instead calculate the average dominance of the 500 nearest neighbors to each group label representation \footnote{We conducted additional analyses by considering 25, 50, 100, 250, and 1000 nearest neighbors, which yielded similar results and can be found in the Supplementary Material}.

As in Section \ref{sec:embed}, we also induced a dominance score directly from a group label's vector representation by adapting the regression-based sentiment prediction from \cite{field2019entity} for word embeddings. More details and results for this technique can be found in the Supplementary Material.

\subsection{Moral Disgust\label{sec:disgust}}

To operationalize \textit{moral disgust} with lexical techniques, we draw inspiration from Moral Foundations theory, which postulates that there are five dimensions of moral intuitions: care, fairness/proportionality, loyalty/ingroup, authority/respect, and sanctity/purity \citep{haidt2007morality}. The negative end of the sanctity/purity dimension corresponds to moral disgust. While we do not directly incorporate Moral Foundations Theory in our framework for dehumanization, we utilize lexicons created by \cite{graham2009liberals} corresponding to each moral foundation. The dictionary for moral disgust includes over thirty words and stems, including \textit{disgust*}, \textit{sin}, \textit{pervert}, and \textit{obscen*} (the asterisks indicate that the dictionary includes all words containing the preceding prefix)\footnote{https://www.moralfoundations.org/othermaterials}.

We opt for a vector approach instead of counting raw frequencies of moral disgust-related words because such words are rare in our news corpus. Furthermore, vectors capture associations with the group label itself, while word counts would not directly capture such associations. Using the word embeddings from Section \ref{sec:embed}, we construct a vector to represent the \textit{concept} of moral disgust by averaging the vectors for all words in the ``Moral Disgust" dictionary, weighted by frequency. This method of creating a vector from the Moral Foundations dictionary resembles that used by \cite{garten2016morality}. We identify implicit associations between a social group and moral disgust by calculating cosine similarity between the group label's vector and the Moral Disgust concept vector, where a higher similarity suggests closer associations between the social group and moral disgust.

\subsection{Vermin as a Dehumanizing Metaphor}

Metaphors comparing humans to vermin have been especially prominent in dehumanizing groups throughout history \citep{haslam2006dehumanization,steuter2010vermin}. Even if a marginalized social group is not directly equated to vermin in the press, this metaphor may be invoked in more subtle ways, such as through the use of verbs that are also associated with vermin (like \textit{scurry} as opposed to the more neutral \textit{hurry}) \citep{marshall2018scurry}. While there is some natural language processing work on the complex task of metaphor detection (e.g. \citeauthor{tsvetkov2014metaphor}, \citeyear{tsvetkov2014metaphor}), these systems cannot easily quantify such indirect associations. 

We thus quantify the metaphorical relationship between a social group and vermin by calculating similarities between these concepts in a distributional semantic vector space. As with \textit{moral disgust}, we create a \textit{Vermin} concept vector by averaging the following vermin words' vectors, weighted by frequency: \textit{vermin}, \textit{rodent(s)}, \textit{rat(s)} \textit{mice}, \textit{cockroaches}, \textit{termite(s)}, \textit{bedbug(s)}, \textit{fleas} \footnote{Largely inspired by https://en.wikipedia.org/wiki/Vermin}. We do not include the singular \textit{mouse} or \textit{flea} because non-vermin senses of those words were more frequent, and word2vec does not account for polysemy. We calculate cosine similarity between each group label and the \textit{Vermin} concept vector, where a high cosine similarity suggests that the group is closely associated with vermin. 

Table \ref{tab:overview} provides an overview of the four elements of dehumanization that we study and the lexical techniques used to quantify them.

\begin{table}[H]

\centering
\begin{tabular}{|c|c|}
\hline

Dehumanization Element & Operationalization \\ \hline \hline
Negative evaluation of target group &  \begin{tabular}[c]{@{}c@{}c@{}}Paragraph-level sentiment analysis\\ Connotation frames of perspective  \\ Word embedding neighbor valence  \end{tabular} \\ \hline 
Denial of agency &  \begin{tabular}[c]{@{}c@{}} Connotation frames of agency \\ Word embedding neighbor dominance  \end{tabular} \\ \hline
Moral disgust & Vector similarity to disgust \\ \hline
Vermin metaphor & Vector similarity to vermin \\ \hline

\end{tabular}
\caption{Overview of linguistic correlates and our operationalizations for four elements of dehumanization.}
\label{tab:overview}
\end{table}

\section{Data}

The data for our case study spans over thirty years of articles from the \textit{New York Times}, from January 1986 to December 2015, and was originally collected by \citeauthor{FastAITrends} \citeyearpar{FastAITrends}. The articles come from all sections of the newspaper, such as ``World", ``New York \& Region", ``Opinion", ``Style", and ``Sports". Our distributional semantic methods rely on all of the available data in order to obtain the most fine-grained understanding of the relationships between words possible. For the other techniques, we extract paragraphs containing any of the following words from a predetermined list of \textbf{LGTBQ terms}: \textit{gay(s), lesbian(s), bisexual(s), homosexual(s), 
transgender(s), transsexual(s), transexual(s), transvestite(s), 
transgendered, asexual, agender, aromantic,  
lgb, lgbt, lgbtq, lgbtqia, 
glbt, lgbtqqia, genderqueer, genderfluid, 
intersex, pansexual}.

Each acronym label is matched insensitive to case and punctuation. Some currently prominent LGBTQ terms, such as \textit{queer} and \textit{trans} are not included in this study, as other senses of these words were more frequent in earlier years. We filter out paragraphs from sections that typically do not pertain to news, such as ``Arts", ``Theater", and ``Movies". While these sections could provide valuable information, we focus on representation of LGBTQ groups in more news-related contexts.

A challenging question when analyzing mass media for subjective attitudes is deciding whose perspective we want to capture: an individual reporter, the institution, or society at large? In this case study, we aim to identify the institution's dehumanizing attitudes towards LGBTQ people. We represent the \textit{New York Times} institution as a combination of the journalists' words in news articles, direct quotes, paraphrases from interviews, and published opinion articles. Therefore, despite our news focus, we include data from ``Opinion" sections; while opinion articles are stylistically different from traditional journalistic reporting due to more overt biases and arguments, these articles are important in constructing the institution's perspective. In addition, we consider all text in each relevant paragraph, including quotes and paraphrases, because they are important to a newspaper's framing of an issue, as particular quotes representing specific stances are intentionally included or excluded from any given article \citep{niculae2015quotus}.

We refer to the remaining subset of the \textit{New York Times} data after filtering as the \textit{LGBTQ corpus}. The \textit{LGBTQ} corpus consists of 93,977 paragraphs and 7.36 million tokens. A large increase in reporting on LGBTQ-related issues has led to a skewed distribution in the amount of data over years, with 1986 containing the least data (1144 paragraphs and 73,549 tokens) and 2012 containing the most (5924 paragraphs and 465,254 tokens).

For all experiments, we also include results for the terms \textit{American} and \textit{Americans}. We include \textit{American(s)} to contrast changes in LGBTQ labels' representation with another social group label. This ensures that the changes we find in dehumanizing language towards LGBTQ groups do not apply uniformly to all social groups, and are thus not merely an artifact of the publication's overall language change. While a natural ``control" variable would be labels such as \textit{straight} or \textit{heterosexual}, these terms only occurred within discussions of LGBTQ communities because they name socially unmarked categories. We also considered comparing LGBTQ labels to \textit{person/people}, but because word embedding-based experiments are sensitive to syntactic forms, we opt for a label that behaves more syntactically similar to \textit{gay} and \textit{homosexual}, particularly with both nominal and adjectival uses. Nevertheless, \textit{American(s)} is by no means a neutral control variable. Because of its in-group status for the \textit{New York Times} (a U.S. institution), we expect our measurements to show that \textit{American(s)} appears in more humanizing contexts than LGBTQ labels; however, we do not expect to find substantial changes in the use of \textit{American(s)} over time.

\begin{figure}
\centering
\includegraphics[width=0.75\textwidth]{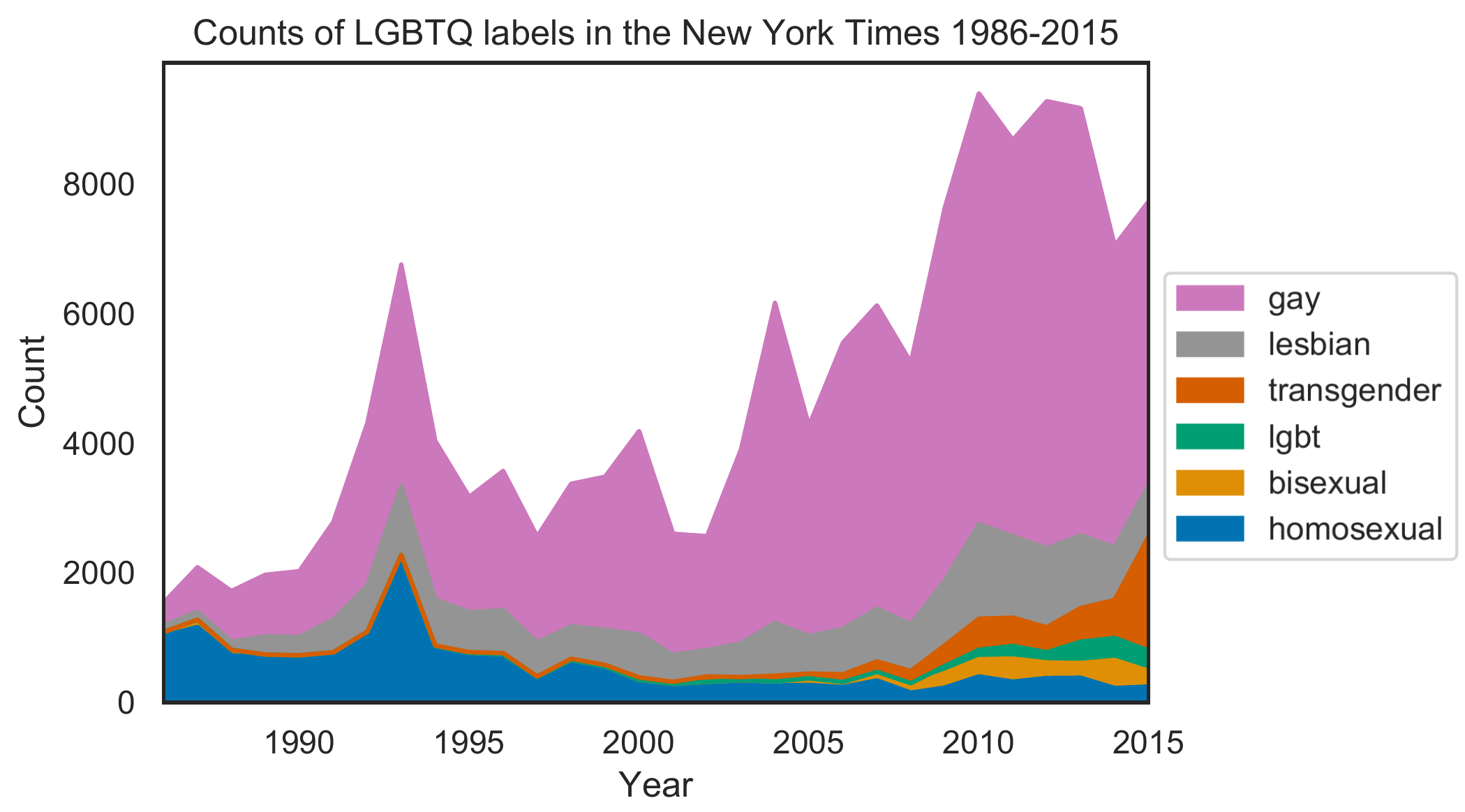}
\caption{Counts for the six most frequent LGBTQ labels in each year of the \textit{New York Times} data.}
\label{fig:lgbt_counts}
\end{figure}

Figure \ref{fig:lgbt_counts} shows the counts of group labels for each year in the \textit{New York Times} from 1986 to 2015. For visualization purposes, only words with a total count greater than 1000 are shown. The relative frequency of \textit{homosexual} decreased substantially over time, while \textit{gay}, \textit{lesbian}, and \textit{bisexual} are more frequent in later years. The terms \textit{lgbt} and \textit{transgender} also emerged after 2000. Counts for all LGBTQ labels can be found in the Supplementary Material.

\section{Results} 

\subsection{Word Embeddings}

Using all of the \textit{New York Times} data, we create word2vec models for each year using the methods described in Section \ref{sec:embed}. Because our computational techniques rely upon these word2vec models, it is useful to gain a sense of how LGBTQ terms are semantically represented within these models. We thus inspect the ten nearest neighbors, or most similar words, to LGBTQ terms in different years. Note that the neighboring words in Tables \ref{tab:all_neighbor} and \ref{tab:neighbor_gay_homosexual} are shown purely for qualitative investigation; our measures for quantifying each dehumanization component incorporate far more information from the word2vec models beyond the top ten neighbors. 

Table \ref{tab:all_neighbor} shows the ten nearest neighbors (by cosine similarity) to our vector representation of all LGBTQ terms, which is the weighted average of the embeddings of all LGBTQ terms considered. For visual convenience, we filter out words occurring fewer than ten times, proper names, as well as other LGBTQ labels and forms of the word \textit{heterosexual}, which are common neighbors for all terms across all years.

Table \ref{tab:all_neighbor} shows that in 1986, LGBTQ groups were most highly associated with words that often convey a sense of sexual deviancy, including \textit{promiscuity}, \textit{promiscuous}, \textit{polygamy}, \textit{bestiality}, and \textit{pornography}. These associations suggest that LGBTQ people were dehumanized to some extent at this time, and their identities were not fully recognized or valued. This shifted by 2000, where we no longer see associations between LGBTQ groups and ideas that evoke moral disgust. Instead, the 2000 vector space shows that LGBTQ people have become more associated with civil rights issues (suggested by \textit{interracial}, \textit{homophobia}, and \textit{discrimination}). The words \textit{ordination} and \textit{ordaining} likely appear due to major controversies that arose at this time about whether LGBTQ people should be permitted to be ordained. We also see some indications of self-identification with the term \textit{openly}. Finally, we see a slight shift towards associations with identity in 2015, with nearby words including \textit{nontransgender}, \textit{closeted}, \textit{equality}, and \textit{sexuality}. Curiously, the word \textit{abortion} is a nearby term for all three years. Perhaps this is because opinions towards abortion and LGBTQ rights seem to be divided along similar partisan lines.

\begin{table}
\centering
\begin{tabular}{|c|c|c|}
\hline
1986 & 2000 & 2015 \\
\hline
sex & interracial & sex \\
premarital & openly & nontransgender \\
sexual & unwed & unmarried \\
abortion & homophobia & interracial \\
promiscuity & premarital & closeted \\
polygamy & ordination & equality \\
promiscuous & nonwhites & couples \\
vigilantism & ordaining & abortion \\
bestiality & discrimination & sexuality \\
pornography & abortion & antiabortion \\
\hline
\end{tabular}
\caption{Nearest words to weighted average of all LGBTQ terms' vectors in 1986, 2000, and 2015}
\label{tab:all_neighbor}
\end{table}

\begin{table}
\centering
\begin{tabular}{|cc|cc|cc|}
\hline
\multicolumn{2}{|c|}{1986} & \multicolumn{2}{|c|}{2000} & \multicolumn{2}{|c|}{2015} \\
Gay & Homosexual & Gay & Homosexual & Gay & Homosexual \\
\hline
homophobia & premarital & interracial & premarital & interracial & premarital \\
women & abortion & openly & openly & sex & sexual \\
feminist & sexual & homophobia & deviant & couples & bestiality \\
vigilante & sex & unwed & interracial & mormons & pedophilia \\
vigilantism & promiscuity & ordination & promiscuity & marriage & adultery \\
suffrage & polygamy & premarital & immoral & closeted & infanticide \\
sexism & anal & abortion & sexual & equality & abhorrent \\
a.c.l.u. & intercourse & antigay & criminalizing & abortion & sex \\
amen & consenting & discrimination & polygamy & unmarried & feticide \\
queer & consensual & marriagelike & consensual & openly & fornication \\ 
\hline
\end{tabular}
\caption{Nearest words to vector representations of \textit{gay} and \textit{homosexual} in 1986, 2000, and 2015}
\label{tab:neighbor_gay_homosexual}
\end{table}

Table \ref{tab:neighbor_gay_homosexual} shows the ten nearest neighboring words to our representations of \textit{gay} and \textit{homosexual} after filtering out proper names, words appearing less than ten times that year, other LGBTQ terms, and forms of \textit{heterosexual}. Table \ref{tab:neighbor_gay_homosexual} reveals variation in social meaning between \textit{gay} and \textit{homosexual} despite denotational similarity, and these differences intensify over time. In 1986, \textit{gay} is associated with terms of discrimination, civil rights and activism, such as \textit{homophobia}, \textit{feminist}, \textit{suffrage}, \textit{sexism}, and \textit{a.c.l.u}. On the other hand, \textit{homosexual} is primarily associated with words related to sexual activity (e.g. \textit{promiscuity}, \textit{anal}, \textit{intercourse}, \textit{consenting}). 

In 1986, this pattern may be due to discussions about sexual transmission of AIDS, but the pejoration of \textit{homosexual} continues over time. While \textit{gay} becomes associated with issues related to marriage equality and identity in 2015, \textit{homosexual} becomes extremely associated with moral disgust and illicit activity, with nearest neighbors including \textit{bestiality}, \textit{pedophilia}, \textit{adultery}, \textit{infanticide}, and \textit{abhorrent.} 

This qualitative analysis of word embedding neighbors reveals significant variation and change in the social meanings associated with LGBTQ group labels, with clear relationships to dehumanizing language. We will now present our quantitative results for measuring each component of dehumanization.

\subsection{Negative Evaluation Towards Target Group}
\subsubsection{Quantitative Results}
\begin{figure}[]
    \centering
    \includegraphics[width=0.6\textwidth]{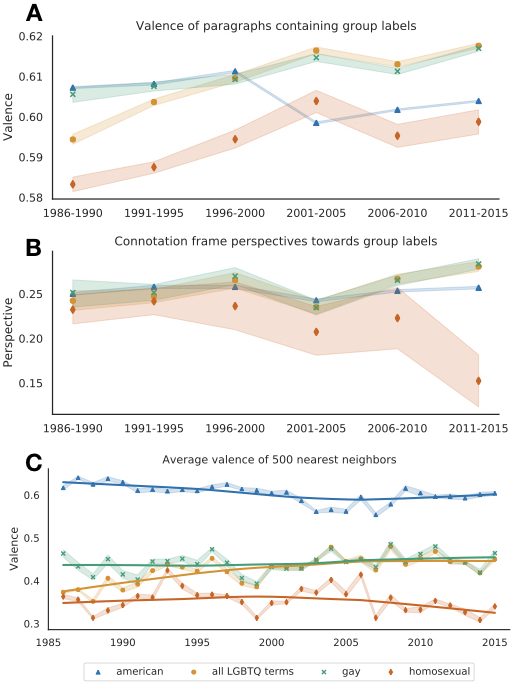}
    \caption{\textbf{(A)} Average paragraph-level valence for paragraphs containing \textit{gay}, \textit{homosexual}, any LGBTQ term, and \textit{American}, grouped into 5-year intervals.  Paragraph-level scores are calculated as the average valence over all words that appear in the NRC VAD Valence Lexicon, which range from 0 (most negative) to 1 (most positive) \citep{vad-acl2018}. Paragraphs containing LGBTQ labels become more positive over time. Paragraphs containing \textit{homosexual} are significantly more negative than those containing other LGBTQ labels. \textbf{(B)} Average connotation frame perspective scores over five-year intervals. Scores are calculated for each subject-verb-object triple containing these group labels as the writer's perspective based on the head verb's entry in the Connotation Frames lexicon \citep{rashkin2016connotation}. \textbf{(C)} Average valence of 500 nearest words to vector representations of \textit{gay}, \textit{homosexual}, \textit{all LGBTQ terms}, and \textit{American}, averaged over 10 word2vec models trained on \textit{New York Times} data from each year. The solid lines are Lowess curves for visualization purposes. Words' valence scores are from the NRC VAD Valence Lexicon. For all plots, the shaded bands represent 95\% confidence intervals.}
    \label{fig:neg_eval}
\end{figure}

\paragraph{Paragraph-level Valence Analysis \label{sec:vad-biases}}

Figure \ref{fig:neg_eval}A shows the average valence for paragraphs containing LGBTQ labels (and \textit{American(s)} for comparison), where a paragraph's valence is simply the average valence over its words (or lemmas) that appear in the NRC VAD Valence Lexicon. The NRC VAD lexicons actually contain several LGBTQ terms, which all have lower than the average valence score of 0.5: \textit{transsexual} (0.264), \textit{homosexual} (0.333), \textit{lesbian} (0.385), \textit{gay} (0.388), and \textit{bisexual} (0.438). These values contrast starkly with more positively-valenced entries in the lexicon, such as \textit{heterosexual} (0.561), \textit{person} (0.646), \textit{human} (0.767), \textit{man} (0.688), and \textit{woman} (0.865). These disparities likely reveal biases among the human annotators whose judgments were used to construct the NRC VAD lexicon \citep{vad-acl2018}. While the lexicon may itself be an interesting artifact of dehumanizing attitudes towards LGBTQ people, we remove these terms before calculating paragraph-level valence scores in order to isolate linguistic signals in the \textit{New York Times} data from annotation biases. Without this preprocessing step, the temporal trends and relative differences between \textit{all LGBTQ terms}, \textit{gay}, and \textit{homosexual} remain roughly the same, but all LGBTQ labels occur in significantly more negative paragraphs than \textit{American}.

Figure \ref{fig:neg_eval}A shows the average paragraph valence. For visualization purposes, we present the results over five-year intervals due to data sparsity in later years for \textit{homosexual} (there were just 208 paragraphs containing \textit{homosexual} in 2014, relative to 3669 containing \textit{gay} in the same year). Analysis of overlapping confidence intervals and Wilcoxon signed-rank tests over the means for each of the thirty years indicates that \textit{gay} and \textit{all LGBTQ terms} occur in significantly more positive paragraphs than \textit{homosexual} ($p < 0.0001$). A linear regression analysis over all years reveals that \textit{all LGBTQ terms}, \textit{gay}, and \textit{homosexual} all significantly increase in paragraph-level valence over time ($p < 0.0001$). However, when considering just the last 15 years, \textit{gay} still significantly increases in paragraph-level valence, while \textit{homosexual} may be trending downward, although this trend does not reach significance in our data ($p = 0.078$). 

The paragraph-level valence analysis shown in Figure \ref{fig:neg_eval}A suggests that LGBTQ groups have become increasingly positively evaluated over time, and thus likely less dehumanized in the \textit{New York Times}. However, the slight downward trend in valence for paragraphs containing \textit{homosexual} between 2001 and 2015 suggests that evaluations of people described as \textit{homosexual} have not improved in the same way as those described by other labels.

Finally, this measurement does not support our initial hypothesis that LGBTQ groups have been more negatively evaluated than \textit{American(s)}, but still reveals that the observed trends for LGBTQ labels are not merely artifacts of changing reporting styles, since paragraphs containing \textit{American(s)} show a very different pattern. Overall, this result demonstrates substantial language change in the \textit{New York Times}'s discussion of LGBTQ people as well as variation in the contexts where different group labels appear, particularly \textit{homosexual}.

\paragraph{Connotation Frames of Perspective}

Figure \ref{fig:neg_eval}B shows the writer's average perspective (valence) towards noun phrases containing either any LGBTQ labels, \textit{gay(s)}, \textit{homosexual(s)}, or the comparison group \textit{American(s)} using the Connotation Frames lexicon \citep{rashkin2016connotation}. The wide variation, particularly for \textit{homosexual}, is likely due to sparsity, as limiting the connotation frames analysis to verbs' immediate subject and direct object noun phrase dependents (consisting of only determiners, adjectives, and nouns) greatly reduced the amount of data for each year; there were only 39 triples for \textit{homosexual} in 2015. We thus show results aggregated over five-year intervals.

As with paragraph-level valence, the writer's perspective towards the label \textit{homosexual} is significantly more negative than towards \textit{gay} ($ p < 0.001$). Linear regression indicates that perspectives towards noun phrases named by any LGBTQ term, \textit{gay}, and \textit{American} have all significantly increased over time ($p < 0.01$). However, the trends are still quite different, as the slopes for \textit{gay} and \textit{all LGBTQ terms} are an order of magnitude greater than \textit{American} 
($m =(1.1\pm0.39)\times10^{-4}$  for \textit{American}, 
$m=(1.4\pm0.18)\times10^{-3}$  for \textit{all LGBTQ terms}, and
$m=(1.1\pm0.22)\times10^{-3}$  for \textit{gay}).
Furthermore, the writer's perspective towards noun phrases containing \textit{homosexual} have significantly decreased over time ($p < 0.0001$). 

Overall, Connotation Frames' perspective scores reveal a similar pattern as the paragraph-level valence analysis, where LGBTQ groups overall appear to be more positively evaluated in the \textit{New York Times} over time. Unlike \textit{gay} and the aggregated \textit{all LGBTQ terms}, the label \textit{homosexual} undergoes pejoration, as \textit{homosexual} becomes increasingly used when (implicitly) expressing negative attitudes towards LGBTQ people.

\paragraph{Word Embedding Neighbor Valence}

Figure \ref{fig:neg_eval}C shows the average valence scores of the 500 nearest neighbors to the vector representations of \textit{gay}, \textit{homosexual}, \textit{all LGBTQ terms}, and \textit{American} for each year. In contrast to our other techniques to quantify \textit{negative evaluations of a target group}, this measurement notably shows that the valence of \textit{American}'s neighboring words is significantly greater than any of the LGBTQ group representations' neighbors every year (Wilcoxon's signed-rank test, $p < 0.0001$), indicating that \textit{American} is used in more positive contexts than LGBTQ terms. Furthermore, all LGBTQ vectors' neighbors have an average valence below the neutral 0.5. The average valence for neighboring words of \textit{gay} and the aggregated \textit{all LGBTQ terms} representation significantly increase over time ($p < 0.0001$), suggesting some increasing humanization in the language used in discussions of LGBTQ people. 

Figure \ref{fig:neg_eval}C also reveals dramatic connotational differences between \textit{gay} and \textit{homosexual}. As shown by non-overlapping confidence intervals and a Wilcoxon signed-rank test, the average valence for \textit{homosexual}'s neighbors is significantly lower than \textit{gay}'s neighbors over all years ($p < 0.0001$). Furthermore, while \textit{gay}'s average neighbor valence increases over time ($p < 0.0001$), \textit{homosexual}'s neighboring words become slightly but significantly more negative over time ($p < 0.001$). Analyzing the valence of the nearest neighbors indicates that \textit{homosexual} has long been used in more negative (and potentially dehumanizing) contexts than \textit{gay}, and that these words' meanings have further diverged as the label \textit{homosexual} has been used in increasingly negative contexts over time.

\subsubsection{Qualitative Analysis}

\paragraph{Paragraph-level Valence Analysis}

How well does paragraph-level valence analysis capture \textit{negative evaluations of a target group}? To facilitate a qualitative evaluation of this technique, we identify several hundred paragraphs with the highest and lowest average valence. Most paragraphs with high valence scores appear to express positive evaluations of LGBTQ individuals, and those with low scores express negative evaluations.

% \usepackage{booktabs}
% \usepackage{graphicx}
\begin{table}
\centering
\resizebox{\textwidth}{!}{%
\begin{tabular}{@{}lllll@{}}
\toprule
Valence & Score & Text & Year & Interpretation \\ \midrule
High & 0.853 & \begin{tabular}[c]{@{}l@{}} All Americans, \textbf{gay} and non-\textbf{gay}, {\color{RoyalBlue}deserve} {\color{blue}respect} and {\color{RoyalBlue}support}\\ for their {\color{blue}families} and basic {\color{blue}freedoms}.\end{tabular} & 2004 & Equality \\ \midrule
High & 0.804 & \begin{tabular}[c]{@{}l@{}}The {\color{RoyalBlue}experience} of the {\color{blue}joy} and {\color{blue}peace} that comes with that — \\ it was a {\color{blue}clear} {\color{RoyalBlue}indication} to me that \textbf{homosexual} {\color{blue}love} was in itself\\ a {\color{blue}good love} and could be a {\color{RoyalBlue}holy} {\color{blue}love},' {\color{RoyalBlue}Father} McNeill said in the {\color{RoyalBlue}film}.\end{tabular} & 2015 & Equality \\ \midrule
High & 0.801 & \begin{tabular}[c]{@{}l@{}}The Straight for {\color{blue}Equality} in {\color{RoyalBlue}Sports} {\color{blue} Award} is {\color{RoyalBlue}given} by PFLAG National,\\ a non-{\color{blue}profit} {\color{RoyalBlue}organization} for {\color{blue}families}, {\color{blue}friends} and {\color{RoyalBlue}allies} of \textbf{gay}, \textbf{lesbian}, \\ \textbf{bisexual} and \textbf{transgender} people.\end{tabular} & 2013 & Advocacy \\ \midrule
High & 0.780 & \begin{tabular}[c]{@{}l@{}}What do you consider the most {\color{blue}interesting} and {\color{RoyalBlue}important} \textbf{LGBT} \\ {\color{RoyalBlue}organizations} {\color{RoyalBlue}working} {\color{RoyalBlue}today} in the {\color{RoyalBlue}city}, with {\color{blue}youth} or more generally? \\ How about more nationally?\end{tabular} & 2010 & Advocacy \\ \midrule \midrule
Low & 0.266 & \begin{tabular}[c]{@{}l@{}}``We {\color{red}kill} the {\color{blue}women}. We {\color{red}kill} the {\color{RoyalBlue}babies}, we {\color{red}kill} the {\color{Rhodamine}blind}. We {\color{red}kill} the {\color{red}cripples}.\\ We {\color{red}kill} them all. We {\color{red}kill} the {\color{Rhodamine}faggot}. We {\color{red}kill} the \textbf{lesbian}... When you {\color{RoyalBlue}get} \\ through {\color{red}killing} them all, go to the {\color{red}goddamn} {\color{red}graveyard} and dig up the {\color{Rhodamine}grave} \\ and {\color{red}kill} them a-{\color{red}goddamn}-{\color{blue}gain} because they didn't {\color{red}die} hard enough."\end{tabular} & 1993 & Direct Quote \\ \midrule
Low & 0.364 & \begin{tabular}[c]{@{}l@{}}A 21-year-old {\color{RoyalBlue}college} student {\color{Rhodamine}pleaded} {\color{red}guilty} yesterday to fatally \\ {\color{Rhodamine}stabbing} a \textbf{gay} man in Queens in what prosecutors termed a {\color{Rhodamine}vicious}\\  {\color{Rhodamine}burst} of anti-\textbf{homosexual} {\color{red}violence}.\end{tabular} & 1991 & Violence \\ \midrule
Low & 0.403 & \begin{tabular}[c]{@{}l@{}}One of his most {\color{Rhodamine}difficult} clients was a \textbf{transsexual} {\color{Rhodamine}prostitute} and \\ {\color{Rhodamine}drug} {\color{red}addict} who was {\color{red}infected} with the AIDS {\color{Rhodamine}virus} and presumably\\ spreading it to her customers and {\color{RoyalBlue}fellow} {\color{red}addicts}.\end{tabular} & 1987 & AIDS \\ \midrule
Low & 0.373 & \begin{tabular}[c]{@{}l@{}} {\color{RoyalBlue}Enabling} {\color{Rhodamine}promiscuity}, indeed! Burroughs Wellcome is as {\color{RoyalBlue}responsible} \\ for the {\color{Rhodamine}reckless} {\color{red}abuse} of amyl nitrate by \textbf{homosexuals} as the \\ manufacturers of {\color{red}narcotic} analgesics are for the {\color{red}horrors} of {\color{Rhodamine}opiate addiction}.\end{tabular} & 1996 & Recklessness \\ \midrule
Low & 0.402 & \begin{tabular}[c]{@{}l@{}}The activists from Africa {\color{Rhodamine}shrugged} with {\color{red}resignation} and sank back {\color{Rhodamine}down} \\ on the benches. The \textbf{gay} Americans absolutely {\color{Rhodamine}exploded} at the {\color{Rhodamine}poor} \\ {\color{RoyalBlue}woman} from the airline.\end{tabular} & 2011 & Recklessness \\ \midrule
Low & 0.397 & \begin{tabular}[c]{@{}l@{}}Homosexuality is {\color{red}forbidden} in Iran. Last year a {\color{RoyalBlue}United} Nations report on \\ {\color{RoyalBlue}human rights} in Iran {\color{RoyalBlue}expressed} {\color{Rhodamine}concern} that \textbf{gays} “face {\color{red}harassment}, \\ {\color{Rhodamine}persecution}, {\color{red}cruel} {\color{red}punishment} and even the {\color{red}death penalty}.\end{tabular} & 2012 & International \\ \bottomrule
\end{tabular}%
}
\caption{Example paragraphs with extremely high and low valence scores, along with an interpretation of the patterns we find. Words with extremely high valence scores (greater than 0.85) appear in {\color{blue}blue}, and somewhat high-valence words (scores between 0.7 and 0.85) appear in {\color{RoyalBlue}light blue}. Words with extremely low valence scores (less than 0.15) appear in {\color{red}red}, and somewhat low-valence words (scores between 0.15 and 0.3) appear in {\color{Rhodamine} pink}.}
\label{tab:para-valence-examples}
\end{table}

Table \ref{tab:para-valence-examples} contain examples with extremely high and low valence. We identify several major themes from these results. Most paragraphs with high valence scores emphasize equal rights, while some focus on the activities of advocacy organizations. On the other end, paragraphs with extremely low valence often focus on violence against LGBTQ people, disease (especially AIDS), and LGBTQ issues internationally. Other themes that emerge in low-valence paragraphs include reports on (and direct quotes from) public figures who dehumanized LGBTQ people and portrayals of LGBTQ people as reckless, irresponsible, and angry.

\begin{table}
\centering
\resizebox{\textwidth}{!}{%
\begin{tabular}{@{}lllll@{}}
\toprule
Valence & Score & Text & Year & Explanation \\ \midrule
High & 0.929 & {\color{blue}Blessing} of \textbf{Homosexuals} & 1990 & Subtitle \\ \midrule
Low & 0.031 & {\color{red}Hate} for Liberals and \textbf{Gays} & 2008 & Subtitle \\ \midrule
High & 0.777 & \begin{tabular}[c]{@{}l@{}}Of the seven in {\color{RoyalBlue}attendance}, only the Rev. Al Sharpton and {\color{RoyalBlue}Representative}\\ Dennis J. Kucinich {\color{blue}supported} \textbf{gay} {\color{RoyalBlue}marriage} unambiguously.\end{tabular} & 2003 & Marriage \\ \midrule
High & 0.765 & \begin{tabular}[c]{@{}l@{}}And I {\color{RoyalBlue}believe} {\color{blue}children} can {\color{RoyalBlue}receive} {\color{blue}love} from \textbf{gay} {\color{RoyalBlue}couples}, but the {\color{RoyalBlue}ideal} is -- \\ and studies {\color{RoyalBlue}have} shown that the {\color{RoyalBlue}ideal} is where a {\color{blue}child} is raised in a married \\ {\color{blue} family} with a man and a {\color{blue}woman}.\end{tabular} & 2005 & \begin{tabular}[c]{@{}l@{}}Marriage\\ Family\end{tabular} \\ \midrule
High & 0.776 & \begin{tabular}[c]{@{}l@{}}Ms. {\color{blue}Bright}, now a {\color{RoyalBlue}college} sophomore, {\color{blue}grew} up in her {\color{blue}mother's home} but \\ regularly visited her \textbf{gay} {\color{RoyalBlue}father}, Lee, in Cartersville, Ga. She {\color{RoyalBlue}remembers} \\ when a {\color{blue}friend} was not {\color{RoyalBlue}allowed} to {\color{RoyalBlue}visit} her {\color{RoyalBlue}father's} {\color{blue}home} because he was gay.\end{tabular} & 1993 & Family \\ \bottomrule
\end{tabular}%
}
\caption{Examples mischaracterized by paragraph-level valence analysis. Words with extremely high valence scores (greater than 0.85) appear in {\color{blue}blue}, and somewhat high-valence words (scores between 0.7 and 0.85) appear in {\color{RoyalBlue}light blue}. Words with extremely low valence scores (less than 0.15) appear in {\color{red}red}, and somewhat low-valence words (scores between 0.15 and 0.3) appear in {\color{Rhodamine} pink}.}
\label{tab:para-valence-flaws}
\end{table}

While this technique accurately captures the valence for many paragraphs, we also identify several shortcomings. Some extreme outliers are extremely short paragraphs, including subtitles within articles which are included as paragraphs in the data. Table \ref{tab:para-valence-flaws} shows several examples that were mischaracterized by our paragraph-level valence analysis technique. In addition, there are several paragraphs with highly positive average valence that actually express negative evaluations of LGBTQ people. The valence of the third paragraph in Table \ref{tab:para-valence-flaws} is skewed by the positive words \textit{supported} and \textit{marriage} even though the paragraph is actually discussing low support for gay marriage. While the fourth paragraph argues that gay couples would be subpar parents relative to straight couples, it uses positive terms such as \textit{love} and \textit{ideal}. Furthermore,  kinship terms tend to be assigned highly positive values in the NRC VAD Valence Lexicon, including \textit{child} and \textit{family}. Similarly, even though the final example describes discrimination based on sexual orientation, the paragraph's average valence is impacted by positive kinship terms such as \textit{father} (0.812) and \textit{mother} (0.931) \footnote{We also conducted paragraph-level sentiment analysis using binary positive vs. negative emotion lexicons such as LIWC \citep{pennebaker2001linguistic}, but found similar quantitative results and no qualitative improvement over the VAD lexicon}.

Overall, our qualitative analysis shows that highly positive valence often accompanies expressions of positive evaluation towards LGBTQ groups, and low valence often accompanies expressions of negative evaluation. However, paragraph-level valence scores are also impacted by specific words cued by various topics; paragraphs about same-sex marriage tend to be more positive because words like \textit{marriage}, \textit{marry}, and \textit{couple} have high valence scores while paragraphs reporting on hate crimes tend to be more negative because they contain low-valence words related to crime, violence, and injury. Furthermore, this method cannot disentangle perspectives within the text; although there are linguistic signals of dehumanization expressed in reports on anti-LGBTQ violence and homophobic speech, these dehumanizing attitudes are not necessarily from the viewpoint of the journalist or the institution. Nevertheless, there could be an overall dehumanizing effect if the media's discussions of a marginalized social group emphasizes such events that harm people. Repeated associations between LGBTQ labels and such negative contexts could potentially contribute to negative evaluations of LGBTQ groups. 

\paragraph{Connotation Frames of Perspective}

To qualitatively analyze how well the connotation frames' lexicon capture \textit{negative evaluation of a target group}, we identify SVO tuples where the verb indicates that the writer has extremely positive or negative perspective towards either the subject or object. The first paragraph in Table \ref{tab:perspective-examples} contains an SVO tuple where the writer has the most negative perspective towards the noun phrases containing a group label. Inside a direct quote, this paragraph uses the phrase \textit{any homosexual act} as the object to the verb \textit{committed}, which has the effect of framing homosexuality as a crime. By deeming gay candidates unworthy of the priesthood, the speaker clearly negatively evaluates LGBTQ people. On the opposite end, many paragraphs labeled as containing extremely positive perspectives towards LGBTQ groups do appear to have positive evaluations of these groups. The second and third paragraphs of Table \ref{tab:perspective-examples} illustrate this, where \textit{the gays} are viewed positively for having \textit{saved} a town, and \textit{gay rights advocates} are \textit{praised} for their work. 

\begin{table}
\centering
\resizebox{\textwidth}{!}{%
\begin{tabular}{@{}lllll@{}}
\toprule
Perspective & Score & Text & SVO & Year \\ \midrule
Negative & -0.83 & \begin{tabular}[c]{@{}l@{}}The most forceful comment came from Cardinal Anthony J. Bevilacqua of \\ Philadelphia, who said his archdiocese screened out gay candidates. `We feel a \\ person who is homosexual-oriented is not a suitable candidate for the priesthood, \\ even if \textbf{he} had never \textbf{committed} \textbf{any homosexual act},' the cardinal said.\end{tabular} & \begin{tabular}[c]{@{}l@{}}S: he\\ V: committed\\ O: any homosexual act\end{tabular} & 2002 \\ \midrule
Positive & +0.80 & \begin{tabular}[c]{@{}l@{}}`Gays are accepted here and respected here,' said Mayor Tony Tarracino. \\ `\textbf{The gays saved a lot} of the oldest parts of town, and they brought in \\ art and culture. They deserve a lot of credit for what Key West is today.'\end{tabular} & \begin{tabular}[c]{@{}l@{}}S: the gays\\ V: saved\\ O: a lot\end{tabular} & 1990 \\ \midrule
Positive & +0.80 & \begin{tabular}[c]{@{}l@{}}In his speech,\textbf{ he praised gay rights advocates} for their hard work and\\ also thanked many elected officials, including his predecessor, \\ Gov. David A. Paterson, and the four Republican state senators who \\ provided the critical votes to pass the marriage bill and whom Mr. Cuomo \\ named one by one to some of the loudest applause of the evening.\end{tabular} & \begin{tabular}[c]{@{}l@{}}S: he\\ V: praised\\ O: gay rights advocates\end{tabular} & 2011 \\ \midrule \midrule

\begin{tabular}[c]{@{}l@{}}Assigned\\ Perspective\end{tabular} & Score & Text & SVO & Year \\ \midrule
Negative & -0.87 & \begin{tabular}[c]{@{}l@{}}Previously, Judge Vaughn Walker, who ruled the ban against\\ same-sex unions unconstitutional in federal court, had said that \\ ProtectMarriage could not appeal his decision to the Ninth Circuit, \\ because they were never able to prove that \textbf{gay marriage harmed them} \\ in any way.\end{tabular} & \begin{tabular}[c]{@{}l@{}}S: gay marriage\\ V: harmed\\ O: them\end{tabular} & 2011 \\ \midrule
Positive & +0.73 & \begin{tabular}[c]{@{}l@{}}Following are excerpts from opinions by the Supreme Court today in\\ its decision that \textbf{the Constitution} does not \textbf{protect private homosexual} \\ \textbf{relations} between consenting adults (...) Justice Stevens wrote a \\ separate dissenting opinion, joined by Justices Brennan and Marshall.\end{tabular} & \begin{tabular}[c]{@{}l@{}}S: the Constitution\\ V: protect\\ O: private homosexual relations\end{tabular} & 1986 \\ \midrule
Positive & +0.70 & \begin{tabular}[c]{@{}l@{}}Do you know there is a Congressional candidate from Missouri who is \\ saying that allowing \textbf{gays} into the military could \textbf{strengthen Al Qaeda}? \\ I’m thinking, how exactly would that work? `They dance better than me, \\ and they know how to accessorize. I’m very, very angry. It’s time for jihad.'\end{tabular} & \begin{tabular}[c]{@{}l@{}}S: gays\\ V: strengthen\\ O: Al Qaeda\end{tabular} & 2010 \\ \bottomrule
\end{tabular}%
}
\caption{Examples of paragraphs where the writer expresses highly positive and negative perspective towards LGBTQ groups, according to the Connotation Frames lexicon. Below the double line are examples of paragraphs where the writer's perspective is mischaracterized by the Connotation Frames lexicon.}
\label{tab:perspective-examples}
\end{table}

However, we found several instances where paragraphs are mislabeled, shown in the bottom half of Table \ref{tab:perspective-examples}. In the fourth paragraph of Table \ref{tab:perspective-examples}, our technique identifies \textit{gay marriage} as the subject of the negative-perspective verb \textit{harmed}, but does not account for the preceding text, which actually contradicts the premise that \textit{gay marriage} causes harm, and thus does not overtly negatively evaluate of LGBTQ groups (although this particular example reveals the difficulty of operationalizing this component because ProtectMarriage groups strongly oppose same-sex marriage and may have negative evaluations of LGBTQ people). The second example similarly shows that this method does not adequately account for various forms of negation, as the positive-perspective verb \textit{protect} is actually negated. The last example in Table \ref{tab:perspective-examples} presents a complex case that is even challenging for qualitative analysis. Our method identifies \textit{gays} as the subject of the verb \textit{strengthen}, even though the subject should be the gerund \textit{allowing gays (into the military)}, and the lexicon's entry for the writer's perspective towards the subject of of \textit{strengthen} is a highly positive 0.7. However, the object of this verb is the terrorist organization \textit{Al Qaeda}; our background knowledge suggests that the capacity to \textit{strengthen} Al Qaeda would reflect negative perspectives. However, this additional context provided by the rest of the paragraph indicates that the writer is being sarcastic and considers the proposition that gays have any impact on strengthening Al Qaeda to be ridiculous. Finally, the writer emphasizes their own stance in opposition to the Missouri congressional candidate by calling upon common stereotypes of gay people being good at dancing and accessorizing.  

Measuring the connotation frames' lexicon perspective scores over verbs' subjects and direct objects cannot leverage as much context as measuring valence over paragraphs using the NRC VAD lexicon labeled for 20,000 words. However, this technique can make more fine-grained distinctions regarding the writer's (and institution's) attitudes directed towards LGBTQ people and is not as dramatically impacted by the emotional valence of the topic discussed. Neither technique can disentangle the journalist's perspective from those expressed by others and simply reported by the journalist. While removing direct quotations may partially address this issue, we deliberately do not remove text from direct quotes or paraphrases. The journalists and newspaper make intentional decisions about what text to include and exclude from quotations, which could still meaningfully represent their perspectives and values \citep{niculae2015quotus}.

\paragraph{Word Embedding Neighbor Valence}

Compared to the previous methods, one limitation of using word embeddings to quantify \textit{negative evaluations of a target group} is that embeddings are not easily interpretable by analyzing a small sample of data. Instead, we assess this technique by identifying LGBTQ terms' nearest neighbors in several outlier years. To facilitate this qualitative analysis, we identify a set of \textit{unique nearest neighbors} for each LGBTQ label in each outlier year, where a word is a unique nearest neighbor for a given LGBTQ term and year if it is not in that term's top 500 nearest neighbors in any other year.

\begin{table}[h]
\centering
\resizebox{\textwidth}{!}{%
\begin{tabular}{@{}lll@{}}
\toprule
Valence & Year & Example \\ \midrule
Low & 1999 & \begin{tabular}[c]{@{}l@{}}Matthew Shepard, a \textbf{gay} college student in Wyoming, had been pistol-whipped and left to die after \\ being tied to a fence on Oct. 7, 1998. Aaron McKinney, who was charged with first-degree murder and \\ other crimes in connection with Mr. Shepard's killing, went on trial Monday, denying that the act was a \\ hate crime, but rather connected to drug use and outrage at a sexual advance he said Mr. Shepard made.\end{tabular} \\ \midrule
Low & 2014 & \begin{tabular}[c]{@{}l@{}}Uganda’s vehement anti-\textbf{gay} movement began in 2009 after a group of American preachers went to \\ Uganda for an anti-\textbf{gay} conference and then worked with Ugandan legislators to draft a bill that called \\ for putting \textbf{gay} people to death. While the bill was being debated, attacks against \textbf{gay} Ugandans began \\ to increase. In early 2011, David Kato, a slight, bespectacled man and one of the country’s most\\ outspoken \textbf{gay} rights activists, was beaten to death with a hammer.\end{tabular} \\ \midrule
Low & 2014 & \begin{tabular}[c]{@{}l@{}}`Hey, @McDonalds: You’re sending \#CheersToSochi while goons wearing Olympic uniforms \\ assault \textbf{LGBT} people,' read one comment last week, from the author and activist Dan Savage.\end{tabular} \\ \midrule
High & 1993 & \begin{tabular}[c]{@{}l@{}}The regulations, which are to take effect Feb. 5, codify the Administration's policy that was worked \\ out as a compromise with the Joints Chiefs of Staff, who had defended the 50-year-old ban, arguing that\\ allowing \textbf{homosexuals} to serve openly would hurt the morale of troops, and thus hurt military readiness.\end{tabular} \\ \bottomrule
\end{tabular}%
}
\caption{Example paragraphs from years where LGBTQ terms' nearest neighbors had exceptionally high and low valence. }
\label{tab:neighbor-valence-examples}
\end{table}

Table \ref{tab:neighbor-valence-examples} contains several example paragraphs that illustrate overarching themes for the outlier years 1993, 1999, and 2014. In 1999, \textit{gay}, \textit{homosexual} and the aggregated representation of \textit{all LGBTQ terms} were all more more closely associated with low-valence words than in almost any other year. We connect this finding to a period of intense reporting in the months following the October 1998 murder of a gay Wyoming college student, Matthew Shepard, which drew national attention to anti-LGBTQ violence. Because LGBTQ labels frequently co-occurred with text about this incident, terms related to Matthew Shepard's case had closer representations to LGBTQ terms in this year. For example, \textit{gay} and \textit{all LGBTQ terms}'s 500 nearest neighbors include \textit{wyoming} in 1999 and \textit{shepard} from the years 1998-2000. Unique nearest neighbors for \textit{gay} in 1999 include other terms that could be connected to this incident, including \textit{homicidal}, \textit{imprisoned}, and \textit{hatred}. Not only was Shepard's murder rooted in the dehumanization of LGBTQ people, but the media's emphasis on the gruesome details of Shepard's death further dehumanized him \citep{ott2002politics}. \citeauthor{ott2002politics} argue that the media's framing of this case actually further  stigmatized LGBTQ people.

Our word embedding neighbor valence measure reveals that the most negative year for \textit{gay} and \textit{all LGBTQ terms} since 1999 was 2014, the second most-recent year of data. We identify several major themes in 2014 that co-occurred with LGBTQ group labels and possibly led to this distributional semantic pattern, primarily reporting on anti-LGBTQ laws and attitudes in Uganda and Russia (particularly in light of the 2014 Winter Olympics in Sochi). The terms \textit{athletes} and \textit{winterolympics} appeared in \textit{gay}'s nearest neighbors in 2014. In addition, the terms \textit{Uganda}, \textit{Ugandan}, and \textit{Mugisha} (a Ugandan LGBT advocate) are among \textit{gay}'s unique nearest 500 neighbors in 2014.

Unlike in 1999 and 2014, LGBTQ terms in 1993 are associated with higher-valence words, especially \textit{homosexual}. \textit{Homosexual}'s unique nearest neighbors in 1993 include the high-valence words \textit{pledge}, \textit{civilian}, \textit{readiness}, and \textit{inclusion}. These words are likely connected with numerous stories in 1993 covering the controversy over whether LGBTQ people should be allowed to serve in the military. 

\subsection{Denial of Agency}
\subsubsection{Quantitative Results}
\begin{figure}
    \centering
    \includegraphics[width=0.6\textwidth]{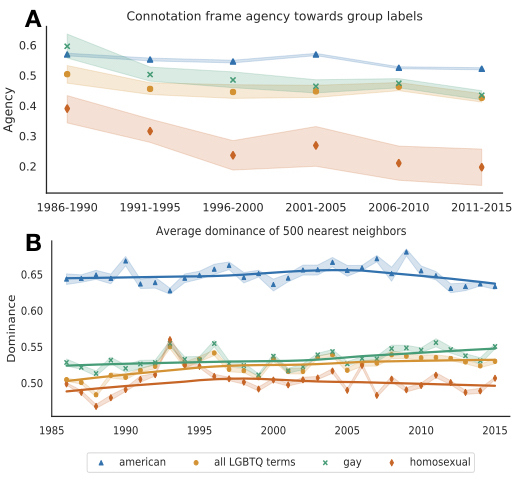}
    \caption{\textbf{(A)} Agency of \textit{gay}, \textit{homosexual}, \textit{all LGBTQ terms}, and \textit{American} using the Connotation Frames lexicon for agency for all subject-verb-object tuples containing each group label \citep{sap2017connotation}, calculated over five-year intervals. An SVO tuple received a score of 1 if the label appears in a positive agency position relative to its head verb and 0 if it does not. \textbf{(B)} Average dominance of 500 nearest words to our representations of \textit{gay}, \textit{homosexual}, \textit{all LGBTQ terms}, and \textit{American}, averaged over 10 word2vec models trained on \textit{New York Times} data for each year. Dominance scores for each word come from the word's entry in the NRC VAD Dominance Lexicon \citep{vad-acl2018}, which range from 0 (least dominance) to 1 (most dominance). For both plots, the shaded bands represent 95\% confidence intervals and the solid lines in (B) are Lowess curves for visualization purposes.}
    \label{fig:denial_agency}
\end{figure}

\paragraph{Connotation Frames of Agency}
Figure \ref{fig:denial_agency}A shows the agency of each group label based on its head verb's entry in the Connotation Frames lexicon for agency  \citep{sap2017connotation}. As in Figure \ref{fig:neg_eval}B, there is large variance due to data sparsity when using the Connotation Frames lexicon, particularly for \textit{homosexual}, which is considerably less frequent than \textit{gay} or other LGBTQ terms in later years. In order to maximize precision when extracting subject-verb pairs, we extract only nouns and their immediate adjectival modifiers, which limits the amount of data. We thus show average agency over five-year intervals.

Wilcoxon signed-rank tests on the means for each group labels over all years indicate that \textit{gay} occurs in contexts with significantly higher agency than \textit{homosexual} ($p < 0.0001$). All four group labels significantly decrease in agency over time according to linear regressions over all 30 years ($p < 0.001$), but the slope for \textit{homosexual} is much greater (
$m=(-7.9\pm1.3)\times10^{-3}$ for \textit{homosexual}, compared to 
$m=(-3.9\pm.55)\times10^{-3}$ for \textit{gay}, and 
$m=(-1.5\pm.46)\times10^{-3}$ for \textit{all LGBTQ terms}). 
Furthermore in the most recent 15 years, \textit{gay} and \textit{all LGBTQ terms} show no significant change ($p = 0.097$ for \textit{gay} and $p = 0.14$ for \textit{all LGBTQ terms}), but \textit{homosexual} still decreases significantly in agency ($p < 0.05$).

Figure \ref{fig:denial_agency}A suggests that LGBTQ groups experience greater denial of agency in the \textit{New York Times} than the institution's in-group identifier \textit{American}. Furthermore, people described as \textit{homosexual} experience even more denial of agency than people who are described as \textit{gay}. Unlike the improving attitudes indicated by our analysis of \textit{negative evaluations of a target group}, it appears that \textit{denial of agency} increased over time for all LGBTQ groups. However, the relatively rapid decrease in agency for \textit{homosexual} is consistent with other results suggesting \textit{homosexual}'s pejoration. 

\paragraph{Word Embedding Neighbor Dominance}

Figure \ref{fig:denial_agency}B shows the average dominance of each group label's 500 nearest neighbors. \textit{American} is significantly associated with greater dominance than \textit{gay}, \textit{homosexual}, and \textit{all LGBTQ terms} (Wilcoxon signed-rank test; $p < 0.0001$), and \textit{gay} has significantly higher dominance than \textit{homosexual} ($p < 0.0001$). While the dominance associated with \textit{gay} and \textit{all LGBTQ terms} significantly increased over time ($p < 0.0001$), the dominance associated with \textit{homosexual} did not significantly change ($p = 0.65$). Furthermore, the average nearest neighbor dominance for \textit{homosexual} decreased in the most recent 15 years ($p < 0.01$).

Even though dominance may more directly encode \textit{power} rather than \textit{agency}, the NRC VAD Dominance Lexicon is useful for operationalizing \textit{denial of agency} because of the close relationship between these concepts. As with Connotation Frames of agency, these results suggest that LGBTQ groups experience greater \textit{denial of agency} than the \textit{New York Times}'s in-group \textit{American}. Both techniques show differences between the labels \textit{gay} and \textit{homosexual}, where \textit{homosexual} is consistently associated with lower agency than \textit{gay} and further decreases over time. However, these two measurements suggest different temporal dynamics for the \textit{denial of agency} of LGBTQ people; Connotation Frames' agency slightly decreases for \textit{all LGBTQ terms} over time, but increases with word embedding neighbor dominance.

\subsubsection{Qualitative Analysis}
\paragraph{Connotation Frames of Agency}

\begin{table}[]
\centering
\resizebox{\textwidth}{!}{%
\begin{tabular}{llll}
\hline
Agency & Text & SVO & Year \\ \hline
High & \begin{tabular}[c]{@{}l@{}}Within the close-knit world of professional childbearers, many of whom\\ share their joys and disillusionments online and in support groups, \\ \textbf{gay couples} have \textbf{developed} \textbf{a reputation} as especially grateful clients...\end{tabular} & \begin{tabular}[c]{@{}l@{}}S: gay couples\\ V: developed\\ O: a reputation\end{tabular} & 2005 \\ \hline
High & \begin{tabular}[c]{@{}l@{}}Tonight, \textbf{the gay rights group} Stonewall Democrats will \textbf{endorse} \textbf{a} \\ \textbf{candidate} for A.G. It’s a relatively big prize in the four-man Democratic \\ primary, given that liberal city voters will have relatively serious sway...\end{tabular} & \begin{tabular}[c]{@{}l@{}}S: the gay rights group\\ V: endorse\\ O: a candidate\end{tabular} & 2006 \\ \hline
Low & \begin{tabular}[c]{@{}l@{}}Nigeria's \textbf{gay men} and lesbians regularly \textbf{face} \textbf{harassment} and arrest, \\ gay activists here say. The criminal code bans acts `against the order \\ of nature,' and imposes sentences of up to 14 years for those convicted...\end{tabular} & \begin{tabular}[c]{@{}l@{}}S: gay men\\ V: face\\ O: harassment\end{tabular} & 2005 \\ \hline
Low & \begin{tabular}[c]{@{}l@{}}Much of the debate among military and civilian officials is now focusing \\ on some version of an approach called `don't ask, don't tell.' (...) But \\ under the `don't tell' element, there would be restrictions on the extent \\ to which \textbf{homosexuals} could \textbf{acknowledge} \textbf{their homosexuality}.\end{tabular} & \begin{tabular}[c]{@{}l@{}}S: homosexuals\\ V: acknowledge\\ O: their homosexuality\end{tabular} & 1993 \\ \hline
\end{tabular}%
}
\caption{Examples where the writer attributes high and low agency towards LGBTQ groups, according to the Connotation Frames lexicon for agency.}
\label{tab:agency-examples}
\end{table}

We qualitatively investigate the labels assigned by this technique for a sample of paragraphs. In general, the binary labels for positive and negative agency seem reasonably accurate, as shown by the first four example in Table \ref{tab:agency-examples}. Verbs that attribute high agency to the subject include \textit{develop} and \textit{endorse}, suggesting that the LGBTQ-aligned subjects are in control and actively making their own decisions. On the other end, LGBTQ people have low agency when they are the subjects of passive verbs such as \textit{face} and \textit{acknowledge}.

The Connotation Frames lexicon for agency seems to be especially accurate for low agency; we could not find counterexamples in our sample where LGBTQ people were portrayed with high agency but labeled with low agency. However, we found several mischaracterizations where LGBTQ people were labeled as having high agency but are not portrayed as agentive or in control of their own actions. Our Connotation Frames technique considers the example below to attribute high agency to LGBTQ people because \textit{homosexual} appears in the subject of the high-agency verb \textit{violate}; however, \textit{homosexual} actually modifies \textit{relationships}, not people themselves. Furthermore, this debate within religion appears to be devoid of input from LGBTQ people and does not portray them as particularly agentive.
\begin{itemize}
    \item At the same time, it underscored a stark division in Judaism over the place of homosexuals in society. Orthodox rabbinical groups believe that \textbf{homosexual relationships violate Jewish law}... (1996)
\end{itemize}

\paragraph{Word Embedding Neighbor Dominance}

Using the VAD Dominance Lexicon to calculate average dominance of each social group label corresponds well to our notion of \textit{denial of agency}. Because \textit{gay}'s nearest neighbors have a much higher average dominance than \textit{homosexual}'s for most years, we compare words that are nearby neighbors for \textit{gay} and not \textit{homosexual} for multiple years' word2vec spaces. Words frequently among the 500 words nearest to \textit{gay} and not \textit{homosexual} include high-agency words such as \textit{activist}, \textit{liberation}, \textit{advocate}, and \textit{advocacy}, which have dominance scores of 0.877, 0.857, 0.818, and 0.731, respectively. Words frequently among \textit{homosexual}'s 500 nearest neighbors and not \textit{gay}'s include low-agency words such as \textit{submissive} (0.173), \textit{degrading} (0.232), \textit{enslavement} (0.302), and \textit{repressed} (0.311). 

We additionally investigate the word2vec models corresponding to several outlier years. \textit{Homosexual}'s neighbors have the highest average dominance in 1993, which is likely due to military-related language in debates surrounding the ``Don't Ask, Don't Tell" legislation.  High-dominance words unique to \textit{homosexual}'s nearest neighbors in 1993 include \textit{forces} (0.886), \textit{military} (0.875), \textit{enforce} (0.836) and \textit{troops} (0.804). \textit{Gay}'s neighbors' in 1999 have the lowest average dominance than any other year, which is likely connected to Matthew Shepard's death and the subsequent outrage; unique neighbors for \textit{gay} in 1999 include \textit{imprisoned} (.302) and \textit{repressed} (0.311).

\subsection{Moral Disgust}

\subsubsection{Quantitative Results} 
\begin{figure}  
\centering
\includegraphics[width=0.75\textwidth]{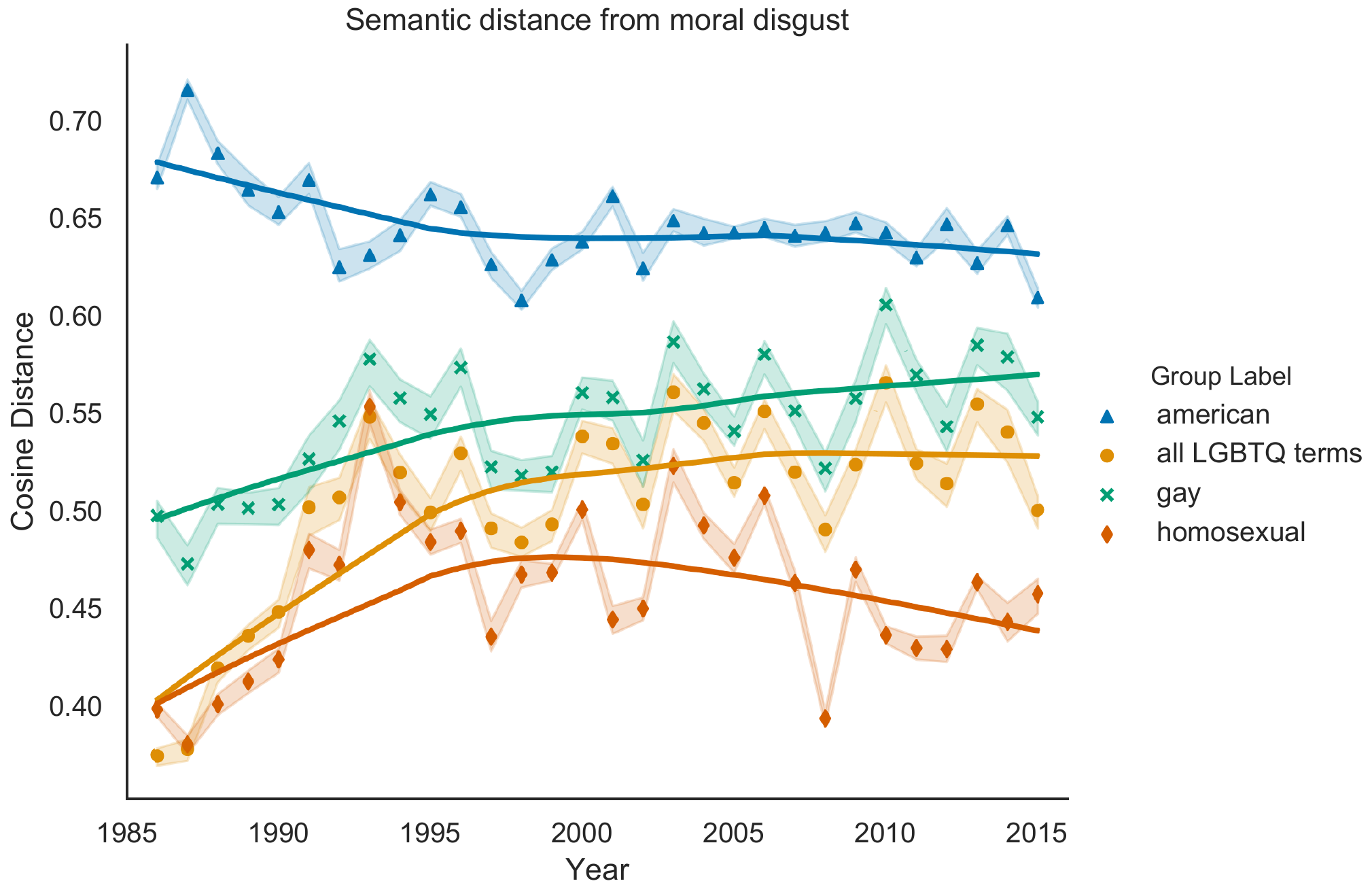}
\caption{Cosine distance between our representations of \textit{gay}, \textit{homosexual}, \textit{all LGBTQ terms}, and \textit{American} and the vector representation of the \textit{Moral Disgust} concept, averaged over 10 word2vec models trained on \textit{New York Times} data for each year. Increases in cosine distance indicate decreases in \textit{Moral Disgust}; possible values range from 0 (most closely associated with Moral Disgust) to 1 (least associated with Moral Disgust). Shaded bands represent 95\% confidence intervals and the solid lines are Lowess curves for visualization purposes. }

\label{fig:disgust_cosine}
\end{figure}

Figure \ref{fig:disgust_cosine} shows the changing relationships between \textit{all LGBTQ terms}, \textit{gay}, \textit{homosexual} and the dehumanizing concept of \textit{Moral Disgust}. Because the cosine distance between \textit{American} and \textit{Moral Disgust} is significantly greater over all years than any LGBTQ representation (Wilcoxon signed-rank test; $p < 0.0001$), \textit{American} is the least associated with \textit{Moral Disgust}. Furthermore, the cosine distance between \textit{gay} and \textit{Moral Disgust} is significantly greater than the distance between \textit{homosexual} and \textit{Moral Disgust} for every year ($p < 0.0001$), indicating that \textit{homosexual} is more closely associated with \textit{Moral Disgust} than \textit{gay} is. Linear regression analyses show that \textit{all LGBTQ terms} and \textit{gay} significantly increase in cosine distance from the \textit{Moral Disgust} vector ($p < 0.0001$), indicated weakening associations between LGBTQ people and moral disgust over time. On the other hand, the distance between \textit{homosexual} and \textit{Moral Disgust} does not change significantly over time ($p = 0.54$), and even decreases after 2000 ($p < 0.05$).

Overall, these measurements of associations between LGBTQ people and \textit{Moral Disgust} are consistent with our other operationalizations of dehumanization. All LGBTQ labels are more closely associated with \textit{Moral Disgust} than the newspaper's in-group term \textit{American}, but these associations weaken over time, suggesting increased humanization. Notably, the term \textit{homosexual} has always been more associated with \textit{Moral Disgust} than the denotationally-similar term \textit{gay}, and \textit{homosexual} actually becomes more closely associated with this dehumanizing concept in recent years.

\subsubsection{Qualitative Analysis}

Our analysis of \textit{homosexual}'s changing semantic neighbors from Table \ref{tab:neighbor_gay_homosexual} has shown that this term has become more associated with immoral concepts, suggesting that moral disgust is a mechanism by which LGBTQ people are dehumanized. Although rarely directly invoked, the connection between LGBTQ people and disgust is supported by the data, such as in the examples shown below, where words belonging to the moral disgust lexicon are in bold. Figure \ref{fig:disgust_cosine} indicates that late 1980s and early 1990s, LGBTQ labels rapidly became more semantically distant from \textit{Moral Disgust}. This likely reflects decreasing attention to the AIDS epidemic, as many disease-related words are included in the moral disgust lexicon.

{\small
\begin{itemize}
    \item Senator Jesse Helms, the North Carolina Republican who has vigorously fought homosexual rights, wants to reduce the amount of Federal money spent on AIDS sufferers, because, he says, it is their ``deliberate, \textbf{disgusting}, revolting conduct" that is responsible for their \textbf{disease}. (1995)
    \item A lawyer named G. Sharp, address unknown, called the cover picture ``utterly \textbf{repulsive}." Donald Ingoglia of Sacramento was equally outraged. ``Showing two smiling gays on the cover illustrates how \textbf{sick} our society has become," he wrote. ``You have my nonlawyer friends falling off their chairs." (1992)
    \item ...Mr. Robison could be harsh -- he yelled in the pulpit and referred to gay men and lesbians as  \textbf{perverts} -- but Mr. Huckabee was a genial ambassador ... (2008)
    \item ...When bishops started telling parishioners that their gay and lesbian siblings were \textbf{sinners}, and that family planning was a grievous wrong, people stopped listening to them -- for good reason. (2013) 
\end{itemize}
}

\subsection{Vermin as a Dehumanizing Metaphor}
\subsubsection{Quantitative Results}
\begin{figure}
\centering
\includegraphics[width=0.75\textwidth]{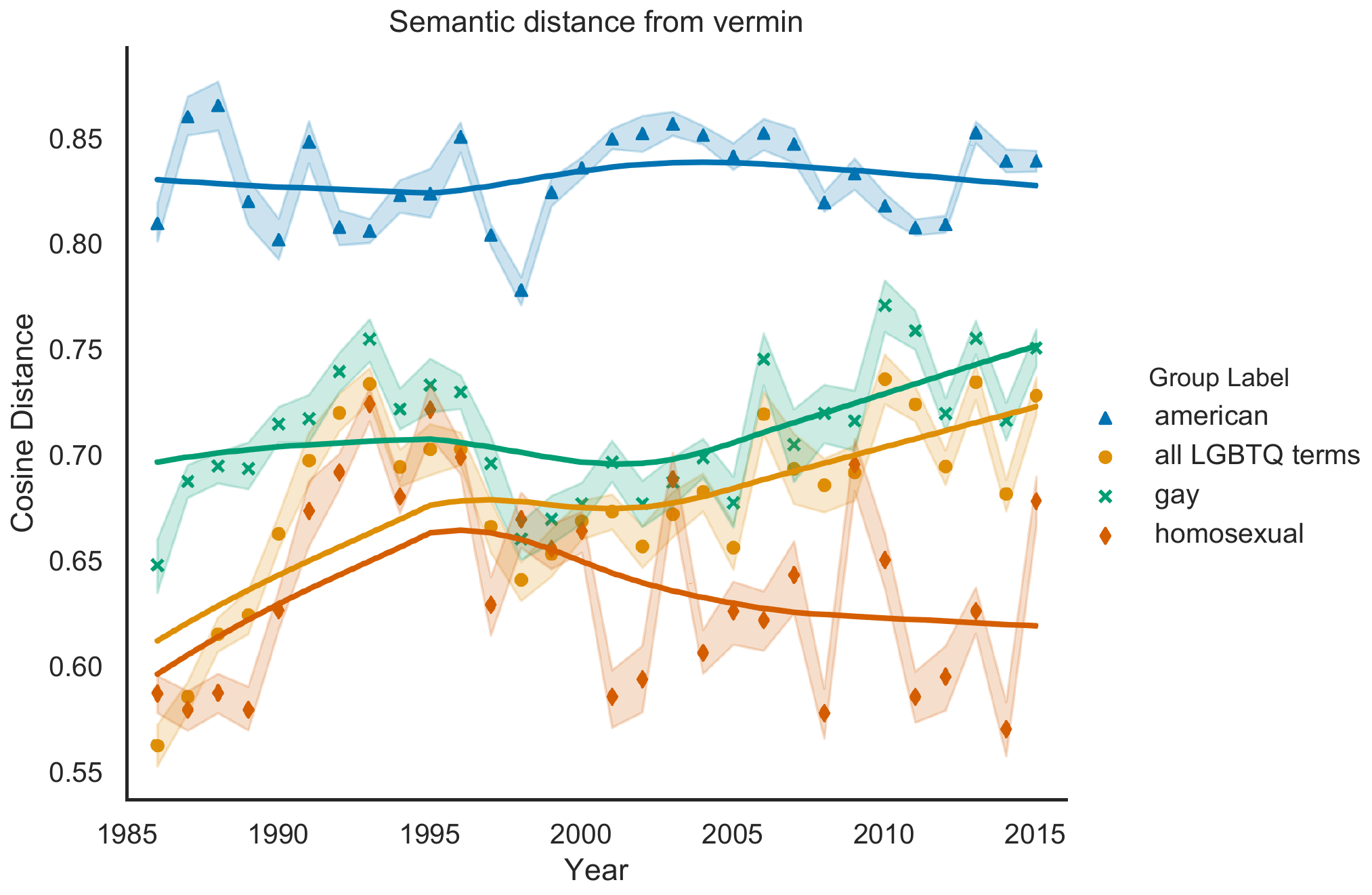}
\caption{Cosine distance between our representations of \textit{gay}, \textit{homosexual}, \textit{all LGBTQ terms}, and \textit{American} and the vector representation of the \textit{Vermin} concept, averaged over 10 word2vec models trained on \textit{New York Times} data for each year. Possible values for cosine distance range from 0 (most closely associated with \textit{Vermin}) to 1 (least associated with \textit{Vermin}). Shaded bands represent 95\% confidence intervals, and the solid lines are Lowess curves for visualization purposes.}
\label{fig:vermin_cosine}
\end{figure}

Figure \ref{fig:vermin_cosine} shows the relationships between LGBTQ labels (and \textit{American}) and the dehumanizing vermin metaphor, quantified as the cosine distance between the labels' word2vec vectors and a \textit{Vermin} concept representation, which is the centroid of multiple vermin-related words. As with \textit{Moral Disgust}, the in-group term \textit{American} is further away from \textit{Vermin} over all years than any LGBTQ term (Wilcoxon signed-rank test; $p < 0.0001$). The cosine distance between \textit{gay} and \textit{Vermin} is also greater than between \textit{homosexual} and \textit{Vermin} ($p < 0.0001$), indicating that \textit{homosexual} is more closely associated with the dehumanizing vermin metaphor than \textit{gay} is. Furthermore, while \textit{all LGBTQ terms} and \textit{gay} become more semantically distant from \textit{Vermin} over time, ( $p < 0.0001$), the association between \textit{Vermin} and \textit{homosexual} does not significantly change over time ($p = 0.13$).

This measure of the implicit \textit{vermin metaphor} reveals similar patterns as the other dehumanization measures. Overall, LGBTQ groups are more associated with vermin than the comparison group \textit{American}, but this association weakens over time, suggesting increased humanization. In addition, \textit{homosexual} has become a more dehumanizing term, with stronger associations with vermin than other LGBTQ labels. 

\subsubsection{Qualitative Analysis}
Metaphors comparing humans to vermin have been especially prominent in dehumanizing groups throughout history \citep{haslam2006dehumanization,steuter2010vermin}. Although no New York Times writers directly compare LGBTQ people to vermin, this metaphor may be invoked in more subtle ways. There are only three paragraphs in the \textit{LGBTQ corpus} that explicitly mention vermin in order to criticize the LGBTQ people-as-vermin metaphor. Nevertheless, these paragraphs point to the existence of this metaphor. 

{\small
\begin{itemize}
    \item Since gay women can't be stigmatized en masse with AIDS, the council had to use real ingenuity to prove that they, too, are vermin at ''much greater risk from one another'' than from gay-bashers ... (1998)
    \item The equating of gay men to vermin is appalling,'' Addessa said from Philadelphia. ``We need to encourage the Eagles and Owens to make a public apology and for the Eagles to publicly discipline Owens. These comments that equate gay men to some inferior life form do real harm, creating a cultural environment which justifies violence against gay and lesbian people. (2004)
    \item In three hours at training camp Tuesday, he hustled vigorously through practice, eagerly signed autographs for visiting military personnel and tried to explain incendiary remarks that appeared in a magazine regarding the sexual orientation of a former teammate  in San Francisco, words that seemed to compare gays to rodents. (2004)
\end{itemize}
}

\section{Human Evaluation of Vector-based Measures}

Our vector-based methods can directly capture associations between LGBTQ people and dehumanizing concepts. However, findings from these methods are difficult to interpret, as discussed in earlier qualitative analysis sections. Furthermore, while the NRC VAD Lexicon and the Connotation Frames Lexicons have been evaluated in prior work \citep{rashkin2016connotation,sap2017connotation,vad-acl2018}, our vector-based methods have not. Thus, we recruit humans from Amazon Mechanical Turk (MTurk) to quantitatively evaluate our four vector-based measures: word embedding neighbor valence (for \textit{negative evaluation of a target group}), word embedding neighbor dominance (for \textit{denial of agency}), semantic distance from the concept of \textit{moral disgust}, and semantic distance from the concept of \textit{vermin}.

Although these four measures rely on vector representations of LGBTQ labels and not individual paragraphs, we use paragraphs as the unit of analysis for our evaluation in order for the task to be feasible for human annotators. In Section \ref{sec:identify_extremes}, we describe how we use our vector-based methods to obtain the most and least dehumanizing paragraphs for each dehumanization component. We discuss the MTurk task design in Section \ref{sec:mturk_design} and results in Section \ref{sec:eval_results}.

\subsection{Identifying the most (de)humanizing paragraphs \label{sec:identify_extremes}}

\subsubsection{Word embedding neighbor valence and dominance \label{sec:filter}}

Our word embedding neighbor valence and dominance methods are proxies for measuring the \textit{negative evaluation of the target group} and \textit{denial of agency} dimensions of dehumanization, respectively. They directly estimate the valence and dominance scores for LGBTQ terms based on NRC VAD entries for each term's semantic neighbors.     

To obtain full paragraphs corresponding to the most and least dehumanizing extremes of \textit{negative evaluation of a target group}, we first train word2vec on the entire \textit{New York Times} dataset using the same hyperparameters as in Section \ref{sec:embed}. Let $N$ be the nearest 500 words to the representation of \textit{all LGBTQ terms} in this vector space, and let $V$ and $D$ be the full NRC Valence and Dominance Lexicons. We define subset lexicons, $V_{s} = N \cap V$ and $D_{s} = N \cap D$; $V_{s}$ and $D_{s}$ are the subsets of the NRC Valence and Dominance Lexicons containing only words that neighbor \textit{all LGBTQ terms}. We  calculate \textit{neighbor valence} scores for each paragraph $P$ as $\frac{1}{\vert P \vert}\Sigma_{w \in P}V_{s}\left[w\right]$, where $\vert P \vert$ is the total number of tokens in $P$ and $V_{s}\left[w\right]$ is the valence score of $w$. Similarly, we calculate \textit{neighbor dominance} scores as $\frac{1}{\vert P \vert}\Sigma_{w \in P}D_{s}\left[w\right]$.

For human evaluation, we consider paragraphs with the highest and lowest scores for \textit{neighbor valence} and \textit{neighbor dominance}. We remove paragraphs containing fewer than 15 or more than 75 words. Because our case study focuses on the words \textit{gay(s)} and \textit{homosexual(s)}, we further restrict our sample to paragraphs containing these terms.

\subsubsection{Moral disgust and vermin metaphor}

We measure implicit associations of LGBTQ groups with \textit{moral disgust} and \textit{vermin} by calculating the cosine distance between LGBTQ terms' vectors and vector representations of \textit{moral disgust} and \textit{vermin}. Thus, we identify paragraphs corresponding to the most and least dehumanizing extremes by comparing the cosine distance between paragraph embeddings and the \textit{Moral Disgust} and \textit{Vermin} concept vectors. We create each paragraph's embedding by calculating the tfidf-weighted average of all words' vectors and removing the first principal component, which improves the quality of sentence and document embeddings \citep{arora2017simple}.

We select the paragraphs that are the closest (most semantically similar) and furthest from the \textit{Moral Disgust} and \textit{Vermin} vectors based on cosine distance. As in Section \ref{sec:filter}, we limit our sample to paragraphs containing between 15 and 75 words and either the term \textit{gay(s)} or \textit{homosexual(s)}.

\subsection{MTurk task design \label{sec:mturk_design}}

As discussed in our qualitative analyses, journalistic text captures numerous perspectives, not only from journalists themselves, but also from people quoted and people or groups described within the text. While our current computational methods do not disambiguate  these perspectives, human evaluation can provide insights into whose perspectives primarily drive our findings about dehumanization. Thus, we manually divide each measure's most and least dehumanizing paragraphs into three categories based on whose views are most prominent: the author, a person quoted or paraphrased, or a person/group mentioned or described within the text. For each measure, our final sample for human evaluation consists of the 20 most humanizing and 20 most dehumanizing paragraphs within each of the three ``viewpoint" categories, yielding 120 paragraphs for each vector-based measure. 

MTurk workers read a paragraph and answered a question about the attitudes of the author, person quoted, or people mentioned/described in the text. Table \ref{tab:eval_examples} shows four examples, the dehumanization component that they correspond to, whether they are ranked high (most dehumanizing) or low (least dehumanizing), the most prominent viewpoint, and the exact question that workers answered. The question depends on which dehumanization component's measure is being evaluated. In addition, we include the actual name of people quoted or mentioned in order to simplify the task. 
Each question is answered with a 5-point Likert scale with endpoints specified in the task. For the \textit{negative evaluation} and \textit{denial of agency} questions, 1 is the most dehumanizing option and 5 is the most humanizing option, but the opposite is the case for  \textit{vermin} and \textit{moral disgust}. As a postprocessing step, we reverse the scale for the latter so higher values always correspond to more humanizing views. 

Three MTurk workers completed each task. Workers were located in the United States, already completed at least 1000 MTurk tasks, and have an approval rate of at least 98\%. Each task took approximately 20-25 seconds and workers were compensated \$0.05. To avoid confusion with multiple question formulations, we published the tasks for each dehumanization component separately.

% \usepackage{booktabs}
% \usepackage{graphicx}
\begin{table}[]
\centering
\resizebox{\textwidth}{!}{%
\begin{tabular}{@{}lllll@{}}
\toprule
Paragraph & Component & Extreme & Viewpoint & Question \\ \midrule
\begin{tabular}[c]{@{}l@{}}Some people think that equality can be achieved by \\ offering gays civil unions in lieu of marriage. Civil \\ unions are not a substitute for marriage. Separate \\ rights are never equal rights.\end{tabular} & \begin{tabular}[c]{@{}l@{}}negative\\ evaluation\end{tabular} & low & author & \begin{tabular}[c]{@{}l@{}}How does the author feel \\ about gay people?\end{tabular} \\ \midrule
\begin{tabular}[c]{@{}l@{}}"I also learned it was possible to be black and gay," \\ Mr. Freeman said. "The first black gay I met, I \\ didn't believe it. I thought you could only be a \\ member of one oppressed minority."\end{tabular} & \begin{tabular}[c]{@{}l@{}}denial of\\ agency\end{tabular} & high & \begin{tabular}[c]{@{}l@{}}person \\ quoted\end{tabular} & \begin{tabular}[c]{@{}l@{}}To what extent does Mr. Freeman\\ think that gay people are able to \\ control their own actions and decisions?\end{tabular} \\ \midrule
\begin{tabular}[c]{@{}l@{}}In a speech exceptional for its deep emotion and \\ sharp message, Ms. Fisher implicitly rebuked those \\ in her party who have regarded the sickness as a \\ self-inflicted plague earned by immoral behavior \\ -- homosexual sex or intravenous drug abuse.\end{tabular} & \begin{tabular}[c]{@{}l@{}}moral \\ disgust\end{tabular} & high & \begin{tabular}[c]{@{}l@{}}person \\ mentioned\end{tabular} & \begin{tabular}[c]{@{}l@{}}To what extent does Ms. Fisher's \\ party consider gay people to be \\ disgusting or repulsive?\end{tabular} \\ \midrule
\begin{tabular}[c]{@{}l@{}}The Supreme Court on Tuesday was deeply divided\\ over one of the great civil rights issues of the age, \\ same-sex marriage. But Justice Anthony M. Kennedy, \\ whose vote is probably crucial, gave gay rights \\ advocates reasons for optimism based on the tone \\ and substance of his questions.\end{tabular} & vermin & low & \begin{tabular}[c]{@{}l@{}}person \\ mentioned\end{tabular} & \begin{tabular}[c]{@{}l@{}}Vermin are animals that carry disease or\\ cause other problems for humans. \\ Examples include rats and cockroaches. \\ To what extent does {[}the author{]} consider \\ gay people to be vermin-like?\end{tabular} \\ \bottomrule
\end{tabular}%
}
\caption{Examples of four paragraphs annotated by MTurk workers, one for each dehumanization component. \textit{Extreme} refers to whether the paragraph is ranked as the most dehumanizing (\textit{high}) or least dehumanizing (\textit{low}) for each measure. \textit{Viewpoint} refers to whose perspective workers are asked to reason about. The question that MTurk workers answer is modified based on both the dehumanization component and the viewpoint.}
\label{tab:eval_examples}
\end{table}

\subsection{Human evaluation results \label{sec:eval_results}}

\begin{figure}
\centering
\includegraphics[width=\textwidth]{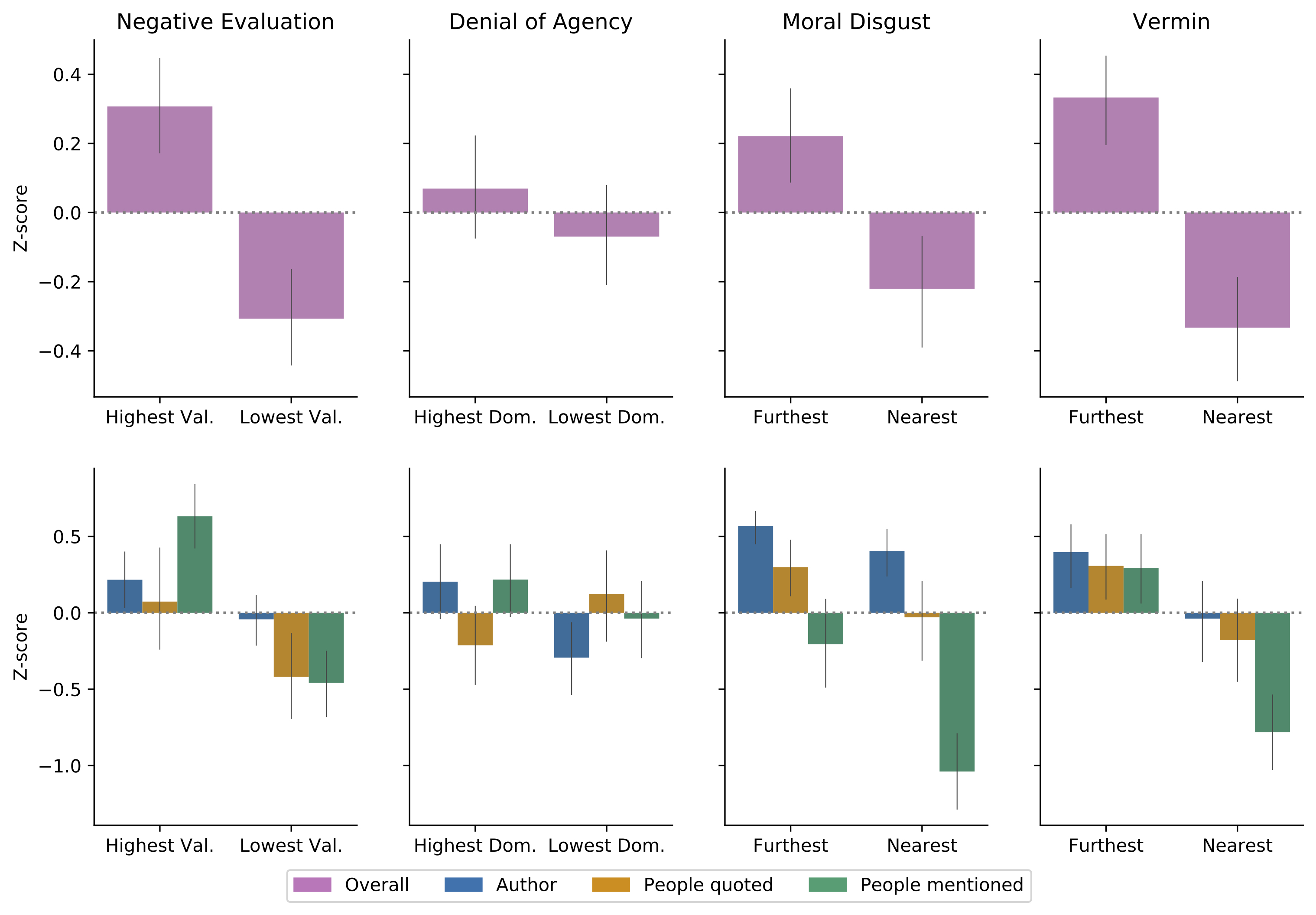}
\caption{Results from human evaluation of our vector-based methods for quantifying \textit{negative evaluation of the target group}, \textit{denial of agency}, \textit{moral disgust}, and \textit{vermin metaphor}. Higher values are more humanizing (more positive evaluation, greater agency, less association with moral disgust or vermin) and lower values are more dehumanizing. The top row shows overall ratings after z-score normalization for each component and the bottom row separates ratings by the viewpoint workers are asked to judge.}
\label{fig:eval_plots}
\end{figure}

The results from the MTurk study, shown in Figure \ref{fig:eval_plots}, largely support our use of vector-based measures. Paragraphs with the highest \textit{neighbor valence} were judged to hold more positive evaluations of gay people ($p < 0.0001$). Paragraphs whose embeddings are nearest to the \textit{Moral Disgust} concept vector are judged to express stronger views of gay people as ``disgusting" or ``repulsive" compared to the furthest paragraphs ($p < 0.0001$). Similarly, paragraphs nearest to \textit{Vermin} concept consider gay people to be more vermin-like than the paragraphs furthest away ($p < 0.0001$). 

The only component that does not follow these expected results is \textit{denial of agency}, where paragraphs with highest and lowest \textit{neighbor dominance} are not judged to be significantly different ($p = 0.19$). This may reflect that using a lexicon for dominance is not a perfect proxy for the more nuanced concept of agency. Another possible explanation is the inherent complexity in measuring \textit{denial of agency}. While the other components are already challenging by requiring an annotator to reason about another person's attitudes towards the target group, assessing \textit{denial of agency} is even more complicated, as it requires an annotator to reason about another person's perceptions of the cognitive capabilities of members of the target group.

The bottom row of Figure \ref{fig:eval_plots} separates the results based on whose viewpoint MTurk workers are asked to reason about: the paragraph's author, the people quoted, or the people mentioned in the text. This reveals a strikingly consistent pattern; the difference between the two extremes is largest when workers are asked about the \textit{people mentioned}, smallest when asked about \textit{the author}, and in-between when asked about \textit{people quoted}. This suggests that dehumanizing representations of LGBTQ people in the \textit{New York Times} may be most driven by descriptions about other people's attitudes, and to a lesser extent, direct quotes and paraphrases.

\section{Discussion}

Our framework for the computational linguistic analysis of dehumanization involves identifying major dimensions of dehumanization from social psychology literature, proposing linguistic correlates for each dimension, and developing robust and interpretable computational methods to quantify these correlates. We apply this framework to study the dehumanization of LGBTQ people in the \textit{New York Times} from 1986 to 2015. We measure four dimensions of dehumanization: \textit{negative evaluations of a target group}, \textit{denial of agency}, \textit{moral disgust}, and (implicit) invocations of \textit{vermin metaphors}. Overall, we find increasingly humanizing descriptions of LGBTQ people over time. LGBTQ people have become more associated with positive emotional language, suggesting that \textit{negative evaluations of the target group} have diminished. LGBTQ people have become more associated with higher-dominance words, suggesting decreasing \textit{denial of agency}, although this finding was not replicated with the verb-centric ``Connotation Frames" measurement. Furthermore, labels for LGBTQ people have moved further away from the concepts of \textit{moral disgust} and \textit{vermin} within distributional semantic representations, suggesting that harmful associations between LGBTQ people and these dehumanizing concepts have weakened over time. 

Despite these trends, the labels \textit{gay} and \textit{homosexual} exhibit strikingly different patterns. \textit{Homosexual} is associated with more negative language than \textit{gay}, suggesting more negative evaluations of people described as \textit{homosexual} than \textit{gay}. \textit{Homosexual} is also associated with greater \textit{denial of agency}, and has stronger connections to \textit{moral disgust} and \textit{vermin} than \textit{gay}. Unlike for other LGBTQ labels, discussions of \textit{homosexual} people have not become more humanizing over time, and several techniques even suggest that \textit{homosexual} has become used in more dehumanizing contexts since 2000. Through its repeated use in these contexts, the use of the word \textit{homosexual} appears to have emerged as an index of more dehumanizing attitudes towards LGBTQ people than other labels. Despite the denotational similarity between \textit{homosexual} and \textit{gay}, our computational techniques tracks the stark divergence in social meanings.  

We restrict our analysis to the lexical level for ease of interpretability, and leveraged a diverse array of existing resources, including the NRC VAD lexicon \citep{vad-acl2018}, Connotation Frames lexicons \citep{rashkin2016connotation,sap2017connotation}, and the Moral Foundations Dictionary \citep{graham2009liberals}. For \textit{negative evaluations of a target group} and \textit{denial of agency}, we propose multiple different techniques that vary in accuracy and interpretability. Word-counting methods are often inaccurate due to their simplicity but their results are easily interpretable, while embedding-based methods suffer the opposite problem. Carefully considering the tradeoff between model quality and interpretability is especially important in work that seeks to characterize complex and sensitive social phenomena such as dehumanization. 

\subsection{Limitations and Future Work}

As the first attempt to computationally analyze dehumanization, this work has many limitations. While we demonstrate how the proposed techniques capture linguistic signals of dehumanization, our qualitative and quantitative evaluation suggest that the findings may be driven more by events and attitudes of people described in the text rather than the journalists' own views. An exciting area of future work could involve developing more sophisticated methods to disambiguate the writer's attitudes, attitudes of people mentioned or quoted, and events, while recognizing that each of these could contribute to the overall representation of marginalized groups in the media. In addition, the present work uses word2vec since all known affective lexicons are type-level, but contextualized embedding-based methods have great potential for more nuanced analyses of dehumanizing language by leveraging token-level representations \citep{devlin2018bert,elmo}. 

Our framework could be expanded to include more insights from dehumanization theory. Beyond the four components discussed in this article, social psychology research has identified other cognitive processes that contribute to dehumanization, including \textit{psychological distancing}, \textit{essentialism} (the perception that the target group has some essence that makes them categorically and fundamentally different), and \textit{denial of subjectivity} (neglect of a target group member's personal feelings and experiences) \citep{rothbart1992category,nussbaum1999sex,graf_nouns,haslam2016recent}. Scholars also differentiate between two forms of dehumanization, \textit{animalistic} (likening humans to animals) and \textit{mechanistic} (likening humans to inanimate objects or machines), which may differ substantially in their linguistic expressions \citep{haslam2006dehumanization}.  

For simplicity and ease of interpretation, we quantify lexical cues of dehumanization. However, our understanding of dehumanizing language would be enriched by considering linguistic features beyond the lexicon. For example, \cite{acton2014pragmatics} has shown that definite plurals in English (e.g. \textit{the gays}) have a unique social and pragmatic effect compared to bare plurals (e.g. \textit{gays}) by packaging individual entities into one monolith and setting this group apart from the speaker. Indexing a speaker's non-membership in the group being discussed creates social distance between the speaker and group \citep{acton2014pragmatics}, which makes it likely that using definite plurals to label marginalized social groups plays an important role in dehumanization. Similarly, examining non-lexical signals could help us capture elements of dehumanization not easily identifiable with lexical resources alone. For example, a group label's word class (e.g. \textit{gay} as a noun or adjective) may have implications for \textit{essentialism}, as adjectives simply name attributes to some entity, while nouns explicitly state the entity's category membership and encapsulates other stereotypes associated with that category \citep{wierzbicka1986s,hall1997red,graf_nouns,palmer2017illegal}. We furthermore believe that incorporating discourse-level analysis, such as examining the role of direct quotes in an article and who is being quoted, could provide informative insights that could address some limitations discussed earlier.  

We support our proposed framework with a case study of LGBTQ representation in the \textit{New York Times}. This case study is limited as an analysis of the dehumanization of LGBTQ people in the media. We only investigate one data source, which does not capture the entirety of media discourse about LGBTQ people. Furthermore, we only study newspaper articles written in (Standard) American English. Future work could focus on cross-linguistic comparisons of dehumanizing language and assess how well our measures generalize to other languages. Finally, the case study focuses on the labels \textit{gay} and \textit{homosexual} due to data availability. As a consequence, we have less understanding about the differences and changes in representations of LGBTQ people who do not identify with these labels. 

The primary aim of this paper is to develop a computational framework for analyzing dehumanizing language towards targeted groups. While our in-depth case study focuses on one particular social group, this framework can be generalized to study dehumanization across a wide variety of social groups, and this could be a fruitful area of future work. For example, Asians have faced increased prejudice and dehumanization since the beginning of the COVID-19 pandemic \citep{van2020using,vidgen2020detecting,ziems2020racism}. Our framework could be applied to understand who dehumanizes these populations in both news and social media, and how the degree of dehumanization changes over time or varies by region. This framework could provide a nuanced view into the shifting nature of dehumanization towards Asians. For example, the ''Asians are good at math" stereotype may have led dehumanization via \textit{denial of agency} or \textit{denial of subjectivity} \citep{shah2019asians}. However, stereotypes of Asians as COVID-19 carriers may have made \textit{moral disgust} and \textit{associations with vermin} more salient mechanisms of dehumanization. In our case study, we use computational measures of dehumanizing language to show how the terms \textit{gay} and \textit{homosexual} have diverged in meaning. This method of demonstrating how denotationally similar items differ in connotation can also generalize to other issues and social groups. For example, we may expect labeling COVID-19 as the \textit{Wuhan Virus} or \textit{Chinese Virus} may be associated with greater dehumanization of Asians than the names \textit{COVID-19} or \textit{SARS-CoV-2} \citep{van2020using,xu2020social}.

\subsection{Ethical Implications}

We hope to draw attention to issues of media representation of marginalized groups and to eventually help make the online world a safer and kinder place. An important part of this mission is to acknowledge the ethical implications and potential issues of our own work. 

The methods that we use to quantify dehumanization are themselves biased and potentially harmful. For example, we show in Section \ref{sec:vad-biases} that the lexicon used to measure valence contains its own anti-LGBTQ biases by considering LGBTQ group labels to be primarily negative/unpleasant. We also train word2vec models on \textit{New York Times} data, which presents biases. Though models trained on biased data are typically concerning due to harmful downstream effects \citep{bolukbasi2016man}, we leverage this data as a  resource for uncovering human biases and understanding \textit{how} biases emerge in the media. 

Another concern of this work in our computational methods to represent human beings. Representing people as sequences of numbers (especially in our vector-based experiments) is inherently dehumanizing. While we hope that this work will humanize and empower marginalized groups, we acknowledge that it can also have effect of perpetuating their dehumanization. 

Other ethical implications of this project appear within our case study. We do not include LGBTQ labels such as \textit{queer} or \textit{trans}, which often had different meanings and were found in unrelated contexts in earlier years. Furthermore, our analysis uses an aggregated representation for LGBTQ people, which unintentionally minimizes the vast diversity of social identities encompassed within this umbrella. We highlight striking temporal changes and differences between \textit{gay} and \textit{homosexual}, which were chosen because these labels were well-represented in all years. However, emphasizing these labels at the expense of others may contribute to the erasure of people who are marginalized even within LGBTQ communities.

\section{Conclusion}

This work is the first known computational linguistic study of dehumanization, and provides contributions to multiple fields. The proposed framework and techniques to quantify salient components of dehumanization can shed light on linguistic variation and change in discourses surrounding marginalized groups. Furthermore, these tools for large-scale analysis have potential to complement smaller-scale psychological studies of dehumanization. Finally, this work has implications for automatically detecting and understanding media bias and abusive language online. 

\section{Data Availability Statement}
We obtained the \textit{New York Times} data through direct communication with one of the creators of the dataset, Ethan Fast \citep{FastAITrends}.

\section{Author Contributions}
JM, YT, and DJ collaborated on the conception and design of the study. JM prepared the data, conducted the case study analysis, and conducted statistical analysis. A first draft of the paper was written by JM. JM, YT, and DJ all read and revised the manuscript, and all authors approved the submitted version.  

\section{Conflict of Interest}
The authors declare that the research was conducted in the absence of any commercial or financial relationships that could be construed as a potential conflict of interest.

\section{Acknowledgements}

We would like to thank Rob Voigt, Chris Potts, Penny Eckert, David Jurgens, Ceren Budak, Anna-Marie Sprenger, and Ismael Mendoza for their insightful contributions and feedback. We would also like to thank our audience at NWAV-46, who provided helpful feedback on an earlier version of this project. Finally, we thank our anonymous reviewers for their thoughtful and valuable comments.

\bibliographystyle{frontiersinSCNS_ENG_HUMS} % for Science, Engineering and Humanities and Social Sciences articles, for Humanities and Social Sciences articles please include page numbers in the in-text citations

\bibliography{test}

\begin{thebibliography}{91}
\providecommand{\natexlab}[1]{#1}
\expandafter\ifx\csname urlstyle\endcsname\relax
  \providecommand{\doi}[1]{doi:\discretionary{}{}{}#1}\else
  \providecommand{\doi}{doi:\discretionary{}{}{}\begingroup
  \urlstyle{rm}\Url}\fi
\providecommand{\selectlanguage}[1]{\relax}
\providecommand{\bibAnnoteFile}[1]{%
  \IfFileExists{#1}{\begin{quotation}\noindent\textsc{Key:} #1\\
  \textsc{Annotation:}\ \input{#1}\end{quotation}}{}}
\providecommand{\bibAnnote}[2]{%
  \begin{quotation}\noindent\textsc{Key:} #1\\
  \textsc{Annotation:}\ #2\end{quotation}}

\bibitem[{Acton(2014)}]{acton2014pragmatics}
Acton, E.~K. (2014).
\newblock \emph{Pragmatics and the social meaning of determiners}.
\newblock Ph.D. thesis, Stanford University
\bibAnnoteFile{acton2014pragmatics}

\bibitem[{Alorainy et~al.(2019)Alorainy, Burnap, Liu, and
  Williams}]{alorainy2019enemy}
Alorainy, W., Burnap, P., Liu, H., and Williams, M.~L. (2019).
\newblock “the enemy among us” detecting cyber hate speech with
  threats-based othering language embeddings.
\newblock \emph{ACM Transactions on the Web (TWEB)} 13, 1--26
\bibAnnoteFile{alorainy2019enemy}

\bibitem[{Arora et~al.(2017)Arora, Liang, and Ma}]{arora2017simple}
Arora, S., Liang, Y., and Ma, T. (2017).
\newblock A simple but tough-to-beat baseline for sentence embeddings
\bibAnnoteFile{arora2017simple}

\bibitem[{Bar-Tal(1990)}]{bar1990causes}
Bar-Tal, D. (1990).
\newblock Causes and consequences of delegitimization: Models of conflict and
  ethnocentrism.
\newblock \emph{Journal of Social issues} 46, 65--81
\bibAnnoteFile{bar1990causes}

\bibitem[{Barnhurst and Mutz(1997)}]{barnhurst1997american}
Barnhurst, K.~G. and Mutz, D. (1997).
\newblock American journalism and the decline in event-centered reporting.
\newblock \emph{Journal of Communication} 47, 27--53
\bibAnnoteFile{barnhurst1997american}

\bibitem[{Baumer et~al.(2015)Baumer, Elovic, Qin, Polletta, and
  Gay}]{baumer2015testing}
Baumer, E., Elovic, E., Qin, Y., Polletta, F., and Gay, G. (2015).
\newblock Testing and comparing computational approaches for identifying the
  language of framing in political news.
\newblock In \emph{Proceedings of the 2015 Conference of the North American
  Chapter of the Association for Computational Linguistics: Human Language
  Technologies}. 1472--1482
\bibAnnoteFile{baumer2015testing}

\bibitem[{Blodgett and O'Connor(2017)}]{blodgett2017racial}
Blodgett, S.~L. and O'Connor, B. (2017).
\newblock Racial disparity in natural language processing: A case study of
  social media african-american english.
\newblock \emph{arXiv preprint arXiv:1707.00061}
\bibAnnoteFile{blodgett2017racial}

\bibitem[{Bolukbasi et~al.(2016)Bolukbasi, Chang, Zou, Saligrama, and
  Kalai}]{bolukbasi2016man}
Bolukbasi, T., Chang, K.-W., Zou, J.~Y., Saligrama, V., and Kalai, A.~T.
  (2016).
\newblock Man is to computer programmer as woman is to homemaker? debiasing
  word embeddings.
\newblock In \emph{Advances in Neural Information Processing Systems}.
  4349--4357
\bibAnnoteFile{bolukbasi2016man}

\bibitem[{Boydstun et~al.(2014)Boydstun, Card, Gross, Resnick, and
  Smith}]{boydstun2014tracking}
Boydstun, A.~E., Card, D., Gross, J., Resnick, P., and Smith, N.~A. (2014).
\newblock Tracking the development of media frames within and across policy
  issues
\bibAnnoteFile{boydstun2014tracking}

\bibitem[{Boydstun et~al.(2013)Boydstun, Gross, Resnik, and
  Smith}]{boydstun2013identifying}
Boydstun, A.~E., Gross, J.~H., Resnik, P., and Smith, N.~A. (2013).
\newblock Identifying media frames and frame dynamics within and across policy
  issues.
\newblock In \emph{New Directions in Analyzing Text as Data Workshop, London}
\bibAnnoteFile{boydstun2013identifying}

\bibitem[{Breitfeller et~al.(2019)Breitfeller, Ahn, Jurgens, and
  Tsvetkov}]{breitfeller-etal-2019-finding}
Breitfeller, L., Ahn, E., Jurgens, D., and Tsvetkov, Y. (2019).
\newblock Finding microaggressions in the wild: A case for locating elusive
  phenomena in social media posts.
\newblock In \emph{Proceedings of the 2019 Conference on Empirical Methods in
  Natural Language Processing and the 9th International Joint Conference on
  Natural Language Processing (EMNLP-IJCNLP)} (Hong Kong, China: Association
  for Computational Linguistics), 1664--1674.
\newblock \doi{10.18653/v1/D19-1176}
\bibAnnoteFile{breitfeller-etal-2019-finding}

\bibitem[{Buckels and Trapnell(2013)}]{buckels2013disgust}
Buckels, E.~E. and Trapnell, P.~D. (2013).
\newblock Disgust facilitates outgroup dehumanization.
\newblock \emph{Group Processes \& Intergroup Relations} 16, 771--780
\bibAnnoteFile{buckels2013disgust}

\bibitem[{Burnap and Williams(2016)}]{burnap2016us}
Burnap, P. and Williams, M.~L. (2016).
\newblock Us and them: identifying cyber hate on twitter across multiple
  protected characteristics.
\newblock \emph{EPJ Data science} 5, 11
\bibAnnoteFile{burnap2016us}

\bibitem[{Caliskan et~al.(2017)Caliskan, Bryson, and
  Narayanan}]{caliskan2017semantics}
Caliskan, A., Bryson, J.~J., and Narayanan, A. (2017).
\newblock Semantics derived automatically from language corpora contain
  human-like biases.
\newblock \emph{Science} 356, 183--186
\bibAnnoteFile{caliskan2017semantics}

\bibitem[{Card et~al.(2015)Card, Boydstun, Gross, Resnik, and
  Smith}]{card2015media}
Card, D., Boydstun, A., Gross, J.~H., Resnik, P., and Smith, N.~A. (2015).
\newblock The media frames corpus: Annotations of frames across issues.
\newblock In \emph{Proceedings of the 53rd Annual Meeting of the Association
  for Computational Linguistics and the 7th International Joint Conference on
  Natural Language Processing (Volume 2: Short Papers)}. 438--444
\bibAnnoteFile{card2015media}

\bibitem[{Demszky et~al.(2019)Demszky, Garg, Voigt, Zou, Shapiro, Gentzkow
  et~al.}]{demszky2019analyzing}
Demszky, D., Garg, N., Voigt, R., Zou, J., Shapiro, J., Gentzkow, M., et~al.
  (2019).
\newblock Analyzing polarization in social media: Method and application to
  tweets on 21 mass shootings.
\newblock In \emph{Proceedings of the 2019 Conference of the North American
  Chapter of the Association for Computational Linguistics: Human Language
  Technologies, Volume 1 (Long and Short Papers)}. 2970--3005
\bibAnnoteFile{demszky2019analyzing}

\bibitem[{Devlin et~al.(2018)Devlin, Chang, Lee, and
  Toutanova}]{devlin2018bert}
Devlin, J., Chang, M.-W., Lee, K., and Toutanova, K. (2018).
\newblock Bert: Pre-training of deep bidirectional transformers for language
  understanding.
\newblock \emph{arXiv preprint arXiv:1810.04805}
\bibAnnoteFile{devlin2018bert}

\bibitem[{Dinakar et~al.(2012)Dinakar, Jones, Havasi, Lieberman, and
  Picard}]{dinakar2012common}
Dinakar, K., Jones, B., Havasi, C., Lieberman, H., and Picard, R. (2012).
\newblock Common sense reasoning for detection, prevention, and mitigation of
  cyberbullying.
\newblock \emph{ACM Transactions on Interactive Intelligent Systems (TiiS)} 2,
  18:1--18:30
\bibAnnoteFile{dinakar2012common}

\bibitem[{Dinu et~al.(2014)Dinu, Lazaridou, and Baroni}]{dinu2014improving}
Dinu, G., Lazaridou, A., and Baroni, M. (2014).
\newblock Improving zero-shot learning by mitigating the hubness problem.
\newblock \emph{arXiv preprint arXiv:1412.6568}
\bibAnnoteFile{dinu2014improving}

\bibitem[{ElSherief et~al.(2018)ElSherief, Kulkarni, Nguyen, Wang, and
  Belding}]{elsherief2018hate}
ElSherief, M., Kulkarni, V., Nguyen, D., Wang, W.~Y., and Belding, E. (2018).
\newblock Hate lingo: A target-based linguistic analysis of hate speech in
  social media.
\newblock In \emph{Twelfth International AAAI Conference on Web and Social
  Media}
\bibAnnoteFile{elsherief2018hate}

\bibitem[{Entman(1993)}]{entman1993framing}
Entman, R.~M. (1993).
\newblock Framing: Toward clarification of a fractured paradigm.
\newblock \emph{Journal of communication} 43, 51--58
\bibAnnoteFile{entman1993framing}

\bibitem[{Esses et~al.(2013)Esses, Medianu, and Lawson}]{esses2013uncertainty}
Esses, V.~M., Medianu, S., and Lawson, A.~S. (2013).
\newblock Uncertainty, threat, and the role of the media in promoting the
  dehumanization of immigrants and refugees.
\newblock \emph{Journal of Social Issues} 69, 518--536
\bibAnnoteFile{esses2013uncertainty}

\bibitem[{Fast and Horvitz(2016)}]{FastAITrends}
Fast, E. and Horvitz, E. (2016).
\newblock Long-term trends in the public perception of artificial intelligence.
\newblock \emph{CoRR} abs/1609.04904
\bibAnnoteFile{FastAITrends}

\bibitem[{Field et~al.(2019)Field, Bhat, and Tsvetkov}]{field2019contextual}
Field, A., Bhat, G., and Tsvetkov, Y. (2019).
\newblock Contextual affective analysis: A case study of people portrayals in
  online\# metoo stories.
\newblock In \emph{Proceedings of the International AAAI Conference on Web and
  Social Media}. vol.~13, 158--169
\bibAnnoteFile{field2019contextual}

\bibitem[{Field et~al.(2018)Field, Kliger, Wintner, Pan, Jurafsky, and
  Tsvetkov}]{field2018framing}
Field, A., Kliger, D., Wintner, S., Pan, J., Jurafsky, D., and Tsvetkov, Y.
  (2018).
\newblock Framing and agenda-setting in russian news: a computational analysis
  of intricate political strategies.
\newblock In \emph{Proceedings of the 2018 Conference on Empirical Methods in
  Natural Language Processing}. 3570--3580
\bibAnnoteFile{field2018framing}

\bibitem[{Field and Tsvetkov(2019)}]{field2019entity}
Field, A. and Tsvetkov, Y. (2019).
\newblock Entity-centric contextual affective analysis.
\newblock In \emph{57th Annual Meeting of the Association for Computational
  Linguistics (ACL 2019)}
\bibAnnoteFile{field2019entity}

\bibitem[{Gallup(2019)}]{gallup2019}
Gallup (2019).
\newblock Gay and lesbian rights.
\newblock http://news.gallup.com/poll/1651/gay-lesbian-rights.aspx
\bibAnnoteFile{gallup2019}

\bibitem[{Garg et~al.(2018)Garg, Schiebinger, Jurafsky, and Zou}]{garg2018word}
Garg, N., Schiebinger, L., Jurafsky, D., and Zou, J. (2018).
\newblock Word embeddings quantify 100 years of gender and ethnic stereotypes.
\newblock \emph{Proceedings of the National Academy of Sciences} 115,
  E3635--E3644
\bibAnnoteFile{garg2018word}

\bibitem[{Garten et~al.(2016)Garten, Boghrati, Hoover, Johnson, and
  Dehghani}]{garten2016morality}
Garten, J., Boghrati, R., Hoover, J., Johnson, K.~M., and Dehghani, M. (2016).
\newblock Morality between the lines: Detecting moral sentiment in text.
\newblock In \emph{Proceedings of IJCAI 2016 workshop on Computational Modeling
  of Attitudes, New York, NY. Retrieved from http://mortezadehghani.
  net/wp-content/uploads/morality-lines-detecting. pdf}
\bibAnnoteFile{garten2016morality}

\bibitem[{Gentzkow and Shapiro(2010)}]{gentzkow2010drives}
Gentzkow, M. and Shapiro, J.~M. (2010).
\newblock What drives media slant? evidence from us daily newspapers.
\newblock \emph{Econometrica} 78, 35--71
\bibAnnoteFile{gentzkow2010drives}

\bibitem[{Goff et~al.(2008)Goff, Eberhardt, Williams, and
  Jackson}]{goff2008not}
Goff, P.~A., Eberhardt, J.~L., Williams, M.~J., and Jackson, M.~C. (2008).
\newblock Not yet human: implicit knowledge, historical dehumanization, and
  contemporary consequences.
\newblock \emph{Journal of personality and social psychology} 94, 292
\bibAnnoteFile{goff2008not}

\bibitem[{Graf et~al.(2013)Graf, Bilewicz, Finell, and Geschke}]{graf_nouns}
Graf, S., Bilewicz, M., Finell, E., and Geschke, D. (2013).
\newblock Nouns cut slices: Effects of linguistic forms on intergroup bias.
\newblock \emph{Journal of Language and Social Psychology} 32, 62--83
\bibAnnoteFile{graf_nouns}

\bibitem[{Graham et~al.(2009)Graham, Haidt, and Nosek}]{graham2009liberals}
Graham, J., Haidt, J., and Nosek, B.~A. (2009).
\newblock Liberals and conservatives rely on different sets of moral
  foundations.
\newblock \emph{Journal of personality and social psychology} 96, 1029
\bibAnnoteFile{graham2009liberals}

\bibitem[{Greene and Resnik(2009)}]{greene2009more}
Greene, S. and Resnik, P. (2009).
\newblock More than words: Syntactic packaging and implicit sentiment.
\newblock In \emph{Proceedings of human language technologies: The 2009 annual
  conference of the north american chapter of the association for computational
  linguistics} (Association for Computational Linguistics), 503--511
\bibAnnoteFile{greene2009more}

\bibitem[{Haidt and Graham(2007)}]{haidt2007morality}
Haidt, J. and Graham, J. (2007).
\newblock When morality opposes justice: Conservatives have moral intuitions
  that liberals may not recognize.
\newblock \emph{Social Justice Research} 20, 98--116
\bibAnnoteFile{haidt2007morality}

\bibitem[{Hall and Moore(1997)}]{hall1997red}
Hall, D.~G. and Moore, C.~E. (1997).
\newblock Red bluebirds and black greenflies: Preschoolers’ understanding of
  the semantics of adjectives and count nouns.
\newblock \emph{Journal of Experimental Child Psychology} 67, 236--267
\bibAnnoteFile{hall1997red}

\bibitem[{Hamilton et~al.(2016)Hamilton, Leskovec, and
  Jurafsky}]{hamilton_semchange}
Hamilton, W.~L., Leskovec, J., and Jurafsky, D. (2016).
\newblock Diachronic word embeddings reveal statistical laws of semantic
  change.
\newblock \emph{CoRR} abs/1605.09096
\bibAnnoteFile{hamilton_semchange}

\bibitem[{Harris and Fiske(2015)}]{harris2015dehumanized}
Harris, L.~T. and Fiske, S.~T. (2015).
\newblock Dehumanized perception.
\newblock \emph{Zeitschrift f{\"u}r Psychologie}
\bibAnnoteFile{harris2015dehumanized}

\bibitem[{Haslam(2006)}]{haslam2006dehumanization}
Haslam, N. (2006).
\newblock Dehumanization: An integrative review.
\newblock \emph{Personality and social psychology review} 10, 252--264
\bibAnnoteFile{haslam2006dehumanization}

\bibitem[{Haslam and Stratemeyer(2016)}]{haslam2016recent}
Haslam, N. and Stratemeyer, M. (2016).
\newblock Recent research on dehumanization.
\newblock \emph{Current Opinion in Psychology} 11, 25--29
\bibAnnoteFile{haslam2016recent}

\bibitem[{Hinds(1977)}]{hinds1977paragraph}
Hinds, J. (1977).
\newblock Paragraph structure and pronominalization.
\newblock \emph{Paper in Linguistics} 10, 77--99
\bibAnnoteFile{hinds1977paragraph}

\bibitem[{Hodson and Costello(2007)}]{hodson2007interpersonal}
Hodson, G. and Costello, K. (2007).
\newblock Interpersonal disgust, ideological orientations, and dehumanization
  as predictors of intergroup attitudes.
\newblock \emph{Psychological Science} 18, 691--698
\bibAnnoteFile{hodson2007interpersonal}

\bibitem[{Katajamaki and Koskela(2006)}]{katajamaki2006rhetorical}
Katajamaki, H. and Koskela, M. (2006).
\newblock The rhetorical structure of editorials in english, swedish and
  finnish business newspapers.
\newblock In \emph{Teoksessa Proceedings of the 5th International Aelfe
  Conference}. 215--19
\bibAnnoteFile{katajamaki2006rhetorical}

\bibitem[{Kiritchenko and Mohammad(2018)}]{kiritchenko2018examining}
Kiritchenko, S. and Mohammad, S. (2018).
\newblock Examining gender and race bias in two hundred sentiment analysis
  systems.
\newblock In \emph{Proceedings of the Seventh Joint Conference on Lexical and
  Computational Semantics}. 43--53
\bibAnnoteFile{kiritchenko2018examining}

\bibitem[{Kteily et~al.(2015)Kteily, Bruneau, Waytz, and
  Cotterill}]{kteily2015ascent}
Kteily, N., Bruneau, E., Waytz, A., and Cotterill, S. (2015).
\newblock The ascent of man: Theoretical and empirical evidence for blatant
  dehumanization.
\newblock \emph{Journal of personality and social psychology} 109, 901
\bibAnnoteFile{kteily2015ascent}

\bibitem[{Kulkarni et~al.(2015)Kulkarni, Al-Rfou, Perozzi, and
  Skiena}]{kulkarni2015statistically}
Kulkarni, V., Al-Rfou, R., Perozzi, B., and Skiena, S. (2015).
\newblock Statistically significant detection of linguistic change.
\newblock In \emph{Proceedings of the 24th International Conference on World
  Wide Web} (International World Wide Web Conferences Steering Committee),
  625--635
\bibAnnoteFile{kulkarni2015statistically}

\bibitem[{Levy and Goldberg(2014)}]{levy2014dependency}
Levy, O. and Goldberg, Y. (2014).
\newblock Dependency-based word embeddings.
\newblock In \emph{Proceedings of the 52nd Annual Meeting of the Association
  for Computational Linguistics (Volume 2: Short Papers)}. 302--308
\bibAnnoteFile{levy2014dependency}

\bibitem[{Manzini et~al.(2019)Manzini, Lim, Tsvetkov, and
  Black}]{manzini2019black}
Manzini, T., Lim, Y.~C., Tsvetkov, Y., and Black, A.~W. (2019).
\newblock Black is to criminal as caucasian is to police: Detecting and
  removing multiclass bias in word embeddings.
\newblock \emph{arXiv preprint arXiv:1904.04047}
\bibAnnoteFile{manzini2019black}

\bibitem[{Marshall and Shapiro(2018)}]{marshall2018scurry}
Marshall, S.~R. and Shapiro, J.~R. (2018).
\newblock When “scurry” vs.“hurry” makes the difference: Vermin
  metaphors, disgust, and anti-immigrant attitudes.
\newblock \emph{Journal of Social Issues} 74, 774--789
\bibAnnoteFile{marshall2018scurry}

\bibitem[{Mikolov et~al.(2013)Mikolov, Sutskever, Chen, Corrado, and
  Dean}]{mikolov2013distributed}
Mikolov, T., Sutskever, I., Chen, K., Corrado, G.~S., and Dean, J. (2013).
\newblock Distributed representations of words and phrases and their
  compositionality.
\newblock In \emph{Advances in neural information processing systems}.
  3111--3119
\bibAnnoteFile{mikolov2013distributed}

\bibitem[{Mohammad(2018)}]{vad-acl2018}
Mohammad, S.~M. (2018).
\newblock Obtaining reliable human ratings of valence, arousal, and dominance
  for 20,000 english words.
\newblock In \emph{Proceedings of The Annual Conference of the Association for
  Computational Linguistics (ACL)} (Melbourne, Australia)
\bibAnnoteFile{vad-acl2018}

\bibitem[{Monroe et~al.(2008)Monroe, Colaresi, and Quinn}]{monroe2008fightin}
Monroe, B.~L., Colaresi, M.~P., and Quinn, K.~M. (2008).
\newblock Fightin'words: Lexical feature selection and evaluation for
  identifying the content of political conflict.
\newblock \emph{Political Analysis} 16, 372--403
\bibAnnoteFile{monroe2008fightin}

\bibitem[{Niculae et~al.(2015)Niculae, Suen, Zhang, Danescu-Niculescu-Mizil,
  and Leskovec}]{niculae2015quotus}
Niculae, V., Suen, C., Zhang, J., Danescu-Niculescu-Mizil, C., and Leskovec, J.
  (2015).
\newblock Quotus: The structure of political media coverage as revealed by
  quoting patterns.
\newblock In \emph{Proceedings of the 24th International Conference on World
  Wide Web} (International World Wide Web Conferences Steering Committee),
  798--808
\bibAnnoteFile{niculae2015quotus}

\bibitem[{Nussbaum(1999)}]{nussbaum1999sex}
Nussbaum, M.~C. (1999).
\newblock \emph{Sex and social justice} (Oxford University Press)
\bibAnnoteFile{nussbaum1999sex}

\bibitem[{Opotow(1990)}]{opotow1990moral}
Opotow, S. (1990).
\newblock Moral exclusion and injustice: An introduction.
\newblock \emph{Journal of social issues} 46, 1--20
\bibAnnoteFile{opotow1990moral}

\bibitem[{Osgood et~al.(1957)Osgood, Suci, and
  Tannenbaum}]{osgood1957measurement}
Osgood, C.~E., Suci, G.~J., and Tannenbaum, P.~H. (1957).
\newblock \emph{The measurement of meaning}.
\newblock 47 (University of Illinois press)
\bibAnnoteFile{osgood1957measurement}

\bibitem[{Ott and Aoki(2002)}]{ott2002politics}
Ott, B.~L. and Aoki, E. (2002).
\newblock The politics of negotiating public tragedy: Media framing of the
  matthew shepard murder.
\newblock \emph{Rhetoric \& Public Affairs} 5, 483--505
\bibAnnoteFile{ott2002politics}

\bibitem[{Palmer et~al.(2017)Palmer, Robinson, and
  Phillips}]{palmer2017illegal}
Palmer, A., Robinson, M., and Phillips, K.~K. (2017).
\newblock Illegal is not a noun: Linguistic form for detection of pejorative
  nominalizations.
\newblock In \emph{Proceedings of the First Workshop on Abusive Language
  Online}. 91--100
\bibAnnoteFile{palmer2017illegal}

\bibitem[{Pennebaker et~al.(2001)Pennebaker, Francis, and
  Booth}]{pennebaker2001linguistic}
Pennebaker, J.~W., Francis, M.~E., and Booth, R.~J. (2001).
\newblock Linguistic inquiry and word count: Liwc 2001.
\newblock \emph{Mahway: Lawrence Erlbaum Associates} 71, 2001
\bibAnnoteFile{pennebaker2001linguistic}

\bibitem[{Peters(2014)}]{peters_2014}
Peters, J.~W. (2014).
\newblock The decline and fall of the ‘h’ word.
\newblock The New York Times
\bibAnnoteFile{peters_2014}

\bibitem[{Peters et~al.(2018)Peters, Neumann, Iyyer, Gardner, Clark, Lee
  et~al.}]{elmo}
Peters, M.~E., Neumann, M., Iyyer, M., Gardner, M., Clark, C., Lee, K., et~al.
  (2018).
\newblock Deep contextualized word representations.
\newblock In \emph{Proc. of NAACL}
\bibAnnoteFile{elmo}

\bibitem[{{Pew Research Center}(2017)}]{PewResearchCenter}
{Pew Research Center} (2017).
\newblock Changing attitudes on gay marriage.
\newblock
  http://www.pewforum.org/fact-sheet/changing-attitudes-on-gay-marriage/
\bibAnnoteFile{PewResearchCenter}

\bibitem[{Pryzant et~al.(2019)Pryzant, Martinez, Dass, Kurohashi, Jurafsky, and
  Yang}]{pryzant2019automatically}
Pryzant, R., Martinez, R.~D., Dass, N., Kurohashi, S., Jurafsky, D., and Yang,
  D. (2019).
\newblock Automatically neutralizing subjective bias in text.
\newblock \emph{arXiv preprint arXiv:1911.09709}
\bibAnnoteFile{pryzant2019automatically}

\bibitem[{Rashkin et~al.(2016)Rashkin, Singh, and
  Choi}]{rashkin2016connotation}
Rashkin, H., Singh, S., and Choi, Y. (2016).
\newblock Connotation frames: A data-driven investigation.
\newblock In \emph{Proceedings of the 54th Annual Meeting of the Association
  for Computational Linguistics (Volume 1: Long Papers)}. 311--321
\bibAnnoteFile{rashkin2016connotation}

\bibitem[{Recasens et~al.(2013)Recasens, Danescu-Niculescu-Mizil, and
  Jurafsky}]{recasens2013linguistic}
Recasens, M., Danescu-Niculescu-Mizil, C., and Jurafsky, D. (2013).
\newblock Linguistic models for analyzing and detecting biased language.
\newblock In \emph{Proceedings of the 51st Annual Meeting of the Association
  for Computational Linguistics (Volume 1: Long Papers)}. 1650--1659
\bibAnnoteFile{recasens2013linguistic}

\bibitem[{Rothbart and Taylor(1992)}]{rothbart1992category}
Rothbart, M. and Taylor, M. (1992).
\newblock Category labels and social reality: Do we view social categories as
  natural kinds?
\bibAnnoteFile{rothbart1992category}

\bibitem[{Rudinger et~al.(2018)Rudinger, Naradowsky, Leonard, and
  Van~Durme}]{rudinger2018gender}
Rudinger, R., Naradowsky, J., Leonard, B., and Van~Durme, B. (2018).
\newblock Gender bias in coreference resolution.
\newblock In \emph{Proceedings of the 2018 Conference of the North American
  Chapter of the Association for Computational Linguistics: Human Language
  Technologies, Volume 2 (Short Papers)}. 8--14
\bibAnnoteFile{rudinger2018gender}

\bibitem[{Russell(1980)}]{russell1980circumplex}
Russell, J.~A. (1980).
\newblock A circumplex model of affect.
\newblock \emph{Journal of personality and social psychology} 39, 1161
\bibAnnoteFile{russell1980circumplex}

\bibitem[{Sap et~al.(2019)Sap, Gabriel, Qin, Jurafsky, Smith, and
  Choi}]{sap2019social}
Sap, M., Gabriel, S., Qin, L., Jurafsky, D., Smith, N.~A., and Choi, Y. (2019).
\newblock Social bias frames: Reasoning about social and power implications of
  language.
\newblock \emph{arXiv preprint arXiv:1911.03891}
\bibAnnoteFile{sap2019social}

\bibitem[{Sap et~al.(2017)Sap, Prasettio, Holtzman, Rashkin, and
  Choi}]{sap2017connotation}
Sap, M., Prasettio, M.~C., Holtzman, A., Rashkin, H., and Choi, Y. (2017).
\newblock Connotation frames of power and agency in modern films.
\newblock In \emph{Proceedings of the 2017 Conference on Empirical Methods in
  Natural Language Processing}. 2329--2334
\bibAnnoteFile{sap2017connotation}

\bibitem[{Schmidt and Wiegand(2017)}]{schmidt2017survey}
Schmidt, A. and Wiegand, M. (2017).
\newblock A survey on hate speech detection using natural language processing.
\newblock In \emph{Proceedings of the Fifth International Workshop on Natural
  Language Processing for Social Media}. 1--10
\bibAnnoteFile{schmidt2017survey}

\bibitem[{Shah(2019)}]{shah2019asians}
Shah, N. (2019).
\newblock “asians are good at math” is not a compliment: Stem success as a
  threat to personhood.
\newblock \emph{Harvard Educational Review} 89, 661--686
\bibAnnoteFile{shah2019asians}

\bibitem[{Sherman and Haidt(2011)}]{sherman2011cuteness}
Sherman, G.~D. and Haidt, J. (2011).
\newblock Cuteness and disgust: the humanizing and dehumanizing effects of
  emotion.
\newblock \emph{Emotion Review} 3, 245--251
\bibAnnoteFile{sherman2011cuteness}

\bibitem[{Shuman(1894)}]{shuman1894steps}
Shuman, E.~L. (1894).
\newblock \emph{Steps into journalism: Helps and hints for young writers}
  (Correspondence School of Journalism)
\bibAnnoteFile{shuman1894steps}

\bibitem[{Silva et~al.(2016)Silva, Mondal, Correa, Benevenuto, and
  Weber}]{silva2016analyzing}
Silva, L., Mondal, M., Correa, D., Benevenuto, F., and Weber, I. (2016).
\newblock Analyzing the targets of hate in online social media.
\newblock In \emph{Tenth International AAAI Conference on Web and Social Media}
  (AAAI), 687--690
\bibAnnoteFile{silva2016analyzing}

\bibitem[{Smith et~al.(2017)Smith, Murib, Motta, Callaghan, and
  Theys}]{smith2017gay}
Smith, B.~A., Murib, Z., Motta, M., Callaghan, T.~H., and Theys, M. (2017).
\newblock “gay” or “homosexual”? the implications of social category
  labels for the structure of mass attitudes.
\newblock \emph{American Politics Research} , 1532673X17706560
\bibAnnoteFile{smith2017gay}

\bibitem[{Soller(2018)}]{soller_2018}
Soller, K. (2018).
\newblock Six times journalists on the paper’s history of covering aids and
  gay issues.
\newblock The New York Times
\bibAnnoteFile{soller_2018}

\bibitem[{Steuter and Wills(2010)}]{steuter2010vermin}
Steuter, E. and Wills, D. (2010).
\newblock ‘the vermin have struck again’: Dehumanizing the enemy in post
  9/11 media representations.
\newblock \emph{Media, War \& Conflict} 3, 152--167
\bibAnnoteFile{steuter2010vermin}

\bibitem[{Sun et~al.(2019)Sun, Gaut, Tang, Huang, ElSherief, Zhao
  et~al.}]{sun2019mitigating}
Sun, T., Gaut, A., Tang, S., Huang, Y., ElSherief, M., Zhao, J., et~al. (2019).
\newblock Mitigating gender bias in natural language processing: Literature
  review.
\newblock In \emph{Proceedings of the 57th Annual Meeting of the Association
  for Computational Linguistics}. 1630--1640
\bibAnnoteFile{sun2019mitigating}

\bibitem[{Tipler and Ruscher(2014)}]{tipler2014agency}
Tipler, C. and Ruscher, J.~B. (2014).
\newblock Agency's role in dehumanization: Non-human metaphors of out-groups.
\newblock \emph{Social and Personality Psychology Compass} 8, 214--228
\bibAnnoteFile{tipler2014agency}

\bibitem[{Tsur et~al.(2015)Tsur, Calacci, and Lazer}]{tsur2015frame}
Tsur, O., Calacci, D., and Lazer, D. (2015).
\newblock A frame of mind: Using statistical models for detection of framing
  and agenda setting campaigns.
\newblock In \emph{Proceedings of the 53rd Annual Meeting of the Association
  for Computational Linguistics and the 7th International Joint Conference on
  Natural Language Processing (Volume 1: Long Papers)}. 1629--1638
\bibAnnoteFile{tsur2015frame}

\bibitem[{Tsvetkov et~al.(2014)Tsvetkov, Boytsov, Gershman, Nyberg, and
  Dyer}]{tsvetkov2014metaphor}
Tsvetkov, Y., Boytsov, L., Gershman, A., Nyberg, E., and Dyer, C. (2014).
\newblock Metaphor detection with cross-lingual model transfer.
\newblock In \emph{Proceedings of the 52nd Annual Meeting of the Association
  for Computational Linguistics (Volume 1: Long Papers)}. 248--258
\bibAnnoteFile{tsvetkov2014metaphor}

\bibitem[{Van~Bavel et~al.(2020)Van~Bavel, Baicker, Boggio, Capraro, Cichocka,
  Cikara et~al.}]{van2020using}
Van~Bavel, J.~J., Baicker, K., Boggio, P.~S., Capraro, V., Cichocka, A.,
  Cikara, M., et~al. (2020).
\newblock Using social and behavioural science to support covid-19 pandemic
  response.
\newblock \emph{Nature Human Behaviour} , 1--12
\bibAnnoteFile{van2020using}

\bibitem[{Vidgen et~al.(2020)Vidgen, Botelho, Broniatowski, Guest, Hall,
  Margetts et~al.}]{vidgen2020detecting}
Vidgen, B., Botelho, A., Broniatowski, D., Guest, E., Hall, M., Margetts, H.,
  et~al. (2020).
\newblock Detecting east asian prejudice on social media.
\newblock \emph{arXiv preprint arXiv:2005.03909}
\bibAnnoteFile{vidgen2020detecting}

\bibitem[{Voigt et~al.(2017)Voigt, Camp, Prabhakaran, Hamilton, Hetey,
  Griffiths et~al.}]{voigt2017language}
Voigt, R., Camp, N.~P., Prabhakaran, V., Hamilton, W.~L., Hetey, R.~C.,
  Griffiths, C.~M., et~al. (2017).
\newblock Language from police body camera footage shows racial disparities in
  officer respect.
\newblock \emph{Proceedings of the National Academy of Sciences} 114,
  6521--6526
\bibAnnoteFile{voigt2017language}

\bibitem[{Wang and Potts(2019)}]{wang2019talkdown}
Wang, Z. and Potts, C. (2019).
\newblock Talkdown: A corpus for condescension detection in context.
\newblock In \emph{Proceedings of the 2019 Conference on Empirical Methods in
  Natural Language Processing and the 9th International Joint Conference on
  Natural Language Processing (EMNLP-IJCNLP)}. 3702--3710
\bibAnnoteFile{wang2019talkdown}

\bibitem[{Wiebe et~al.(2004)Wiebe, Wilson, Bruce, Bell, and
  Martin}]{wiebe2004learning}
Wiebe, J., Wilson, T., Bruce, R., Bell, M., and Martin, M. (2004).
\newblock Learning subjective language.
\newblock \emph{Computational linguistics} 30, 277--308
\bibAnnoteFile{wiebe2004learning}

\bibitem[{Wierzbicka(1986)}]{wierzbicka1986s}
Wierzbicka, A. (1986).
\newblock What's in a noun?(or: How do nouns differ in meaning from
  adjectives?).
\newblock \emph{Studies in Language. International Journal sponsored by the
  Foundation “Foundations of Language”} 10, 353--389
\bibAnnoteFile{wierzbicka1986s}

\bibitem[{Xu and Liu(2020)}]{xu2020social}
Xu, C. and Liu, M.~Y. (2020).
\newblock Social cost with no political gain: The" chinese virus" effect
\bibAnnoteFile{xu2020social}

\bibitem[{Zhao et~al.(2018)Zhao, Wang, Yatskar, Ordonez, and
  Chang}]{zhao2018gender}
Zhao, J., Wang, T., Yatskar, M., Ordonez, V., and Chang, K.-W. (2018).
\newblock Gender bias in coreference resolution: Evaluation and debiasing
  methods.
\newblock In \emph{Proceedings of the 2018 Conference of the North American
  Chapter of the Association for Computational Linguistics: Human Language
  Technologies}. vol.~2
\bibAnnoteFile{zhao2018gender}

\bibitem[{Ziems et~al.(2020)Ziems, He, Soni, and Kumar}]{ziems2020racism}
Ziems, C., He, B., Soni, S., and Kumar, S. (2020).
\newblock Racism is a virus: Anti-asian hate and counterhate in social media
  during the covid-19 crisis.
\newblock \emph{arXiv preprint arXiv:2005.12423}
\bibAnnoteFile{ziems2020racism}

\end{thebibliography}

\newpage

\section*{Supplementary Material}

% \usepackage{booktabs}
% \usepackage{graphicx}
\begin{table}[h]
\centering
%\resizebox{4cm}{!}{%
%\begin{tabular}{@{}ll@{}}
\begin{tabular}{llllllll}
\toprule
Label & Count & Label & Count & Label & Count & Label & Count\\ \midrule \midrule
gay(s) & 96977 & lgbt & 1783 & lgbtq & 129 & agender & 10\\ \midrule 
lesbian(s) & 20233 & transvestite(s) & 625 & glbt & 68 & aromantic & 5\\ \midrule
homosexual(s) & 16638 & tran(s)sexual(s) & 627 & genderqueer & 51 & lgbtqia & 4\\ \midrule
transgender(s/ed) & 6066 & asexual & 255 & lgb & 29 & genderfluid & 0\\ \midrule
bisexual(s) & 3464 & intersex & 210 & pansexual & 22 & lgbtqqia & 0 \\ \midrule
\end{tabular}%
 %}
\caption{Overall counts for all LGBTQ terms in the New York Times from 1986-2015. Each label includes its morphological and orthographic variants.}
\label{tab:term_counts}
\end{table}

\newpage

\subsection*{Nearest neighbor valence with different thresholds}
\begin{figure}[h]
    \centering
    \includegraphics[width=\textwidth]{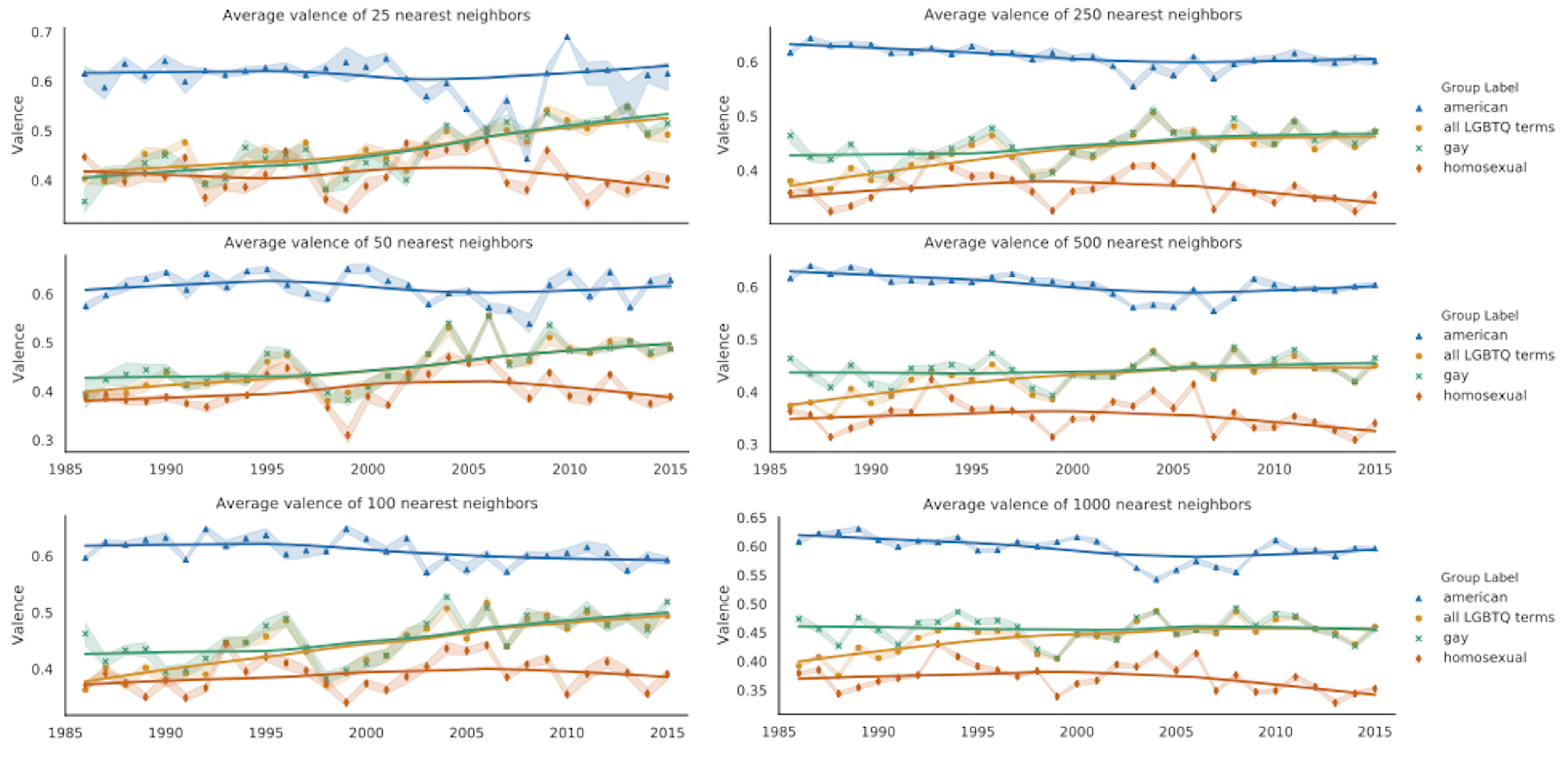}
    \caption{Average valence of 25, 50, 100, 250, 500 and 1000 nearest words to vector representations of \textit{gay}, \textit{homosexual}, \textit{all LGBTQ terms}, and \textit{American}, averaged over 10 word2vec models trained on \textit{New York Times} data from each year. The solid lines are Lowess curves for visualization purposes. Words' valence scores are from the NRC VAD Valence Lexicon. For all plots, the shaded bands represent 95\% confidence intervals.}
\end{figure}

\newpage

\subsection*{Nearest neighbor dominance with different thresholds}
\begin{figure}[h]
    \centering
    \includegraphics[width=\textwidth]{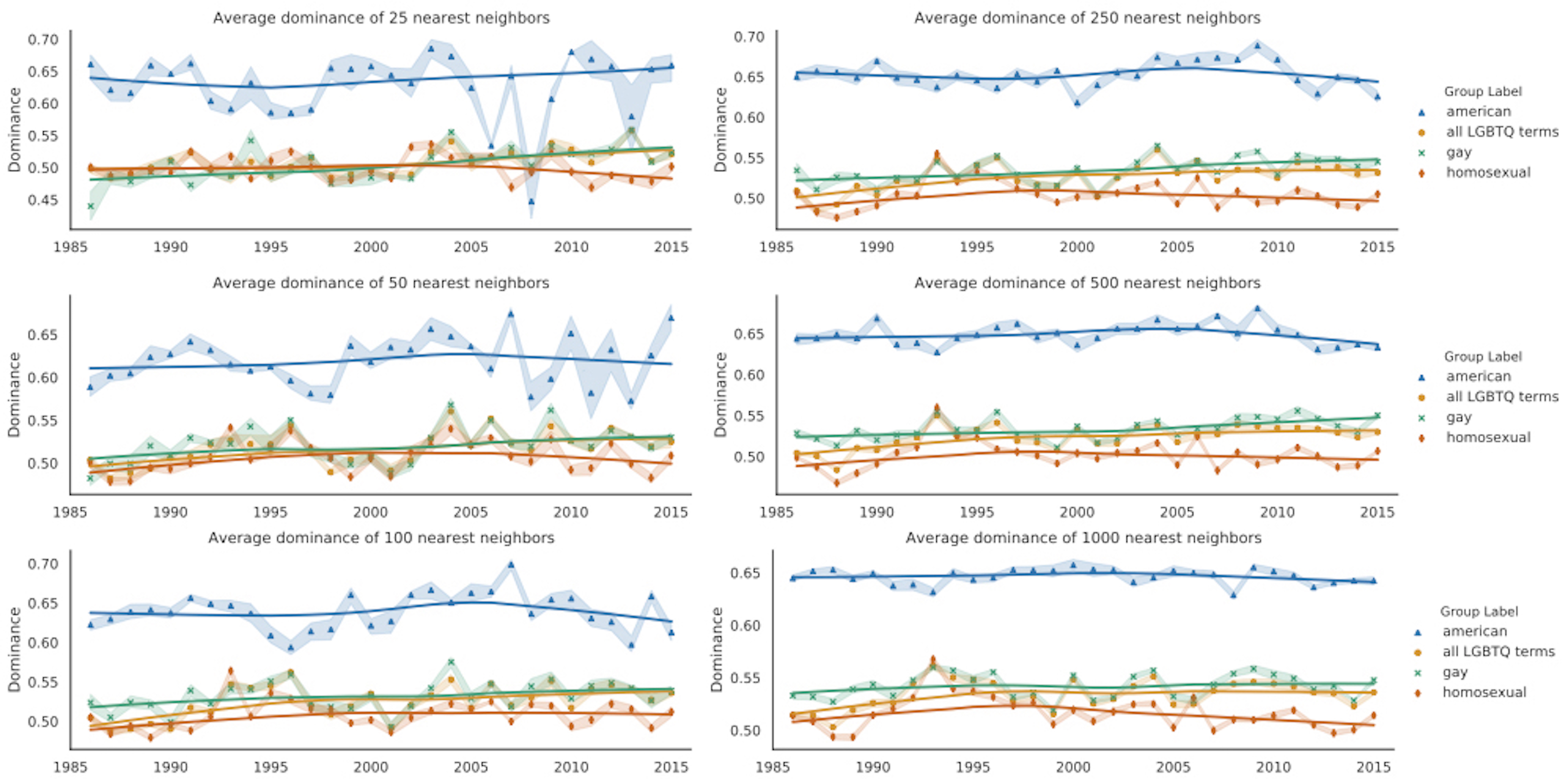}
     \caption{Average dominance of 25, 50, 100, 250, 500 and 1000 nearest words to vector representations of \textit{gay}, \textit{homosexual}, \textit{all LGBTQ terms}, and \textit{American}, averaged over 10 word2vec models trained on \textit{New York Times} data from each year. The solid lines are Lowess curves for visualization purposes. Words' dominance scores are from the NRC VAD Dominance Lexicon. For all plots, the shaded bands represent 95\% confidence intervals.}

\end{figure}

\subsection*{Valence prediction results}

In addition to quantifying the \textit{negative evaluation of a target group} by calculating the average valence score of a group label's vector representation's nearest neighbors according to the NRC VAD lexicon, we also directly induced a valence score from the vector representation itself. We use the zero-centered, normalized, word embeddings created for each year as features in a regression model to directly predict valence (Field et al., 2019). Specifically, we train ridge regression models for each year, where each year's Word2Vec representation for words from the NRC VAD lexicon are features and the lexicon's valence scores are labels. 85\% of words from the VAD lexicon were kept as the training set, and the remaining 15\% was used as a test set to evaluate performance. We then use this set of regression models to predict a group label's valence from its vector representation.   

\begin{figure}[]
\centering
\includegraphics[width=0.75\textwidth]{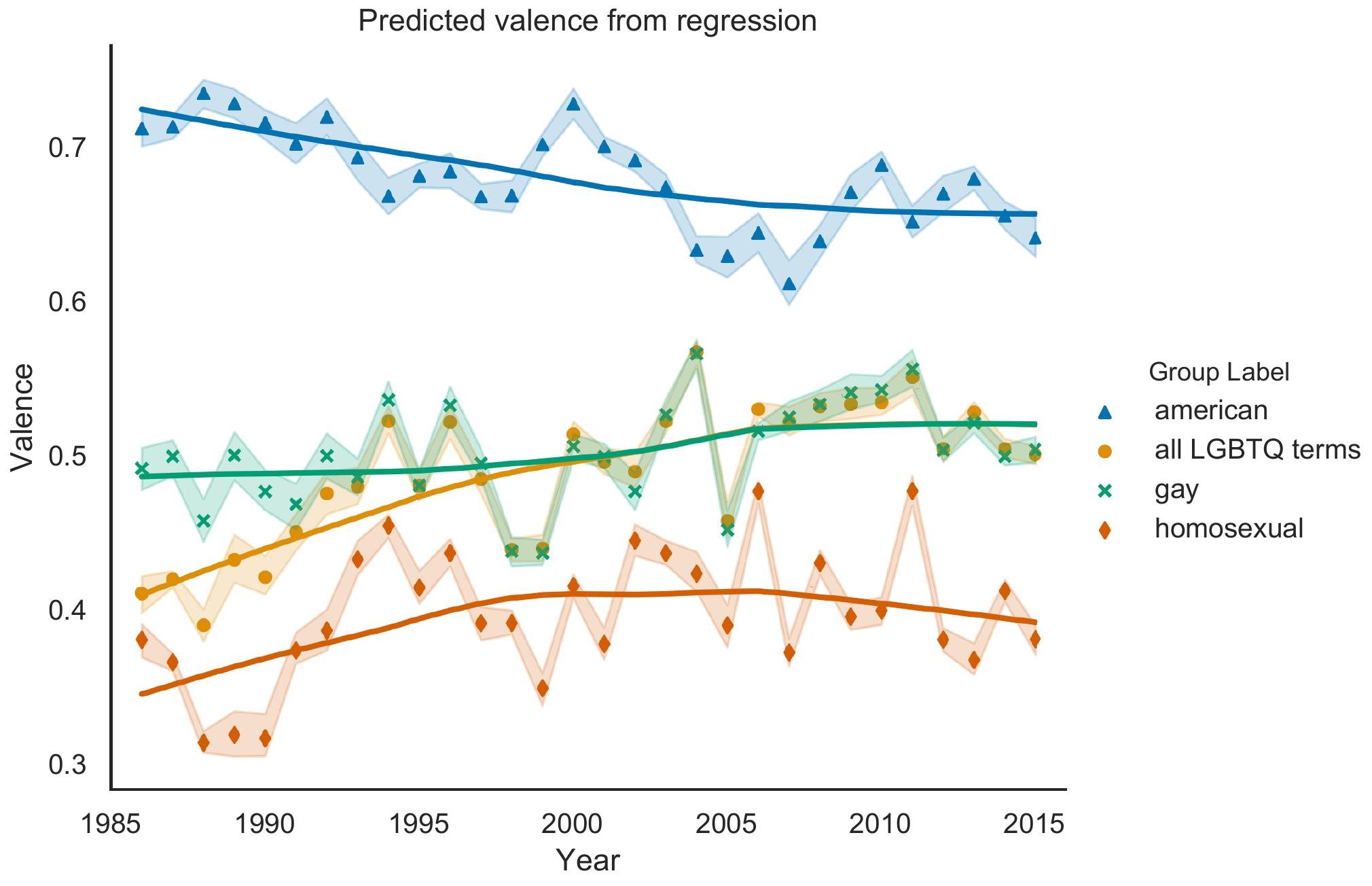}
\caption{Predicted valence of our representations of \textit{gay}, \textit{homosexual},\textit{all LGBTQ terms}, and \textit{American} directly induced by fitting ridge regression models to the NRC VAD Valence Lexicon with the lexicon's words' Word2Vec vectors as features for each year. Results are averaged over 10 Word2Vec models trained over each year's data. Shaded bands represent 95\% confidence intervals and the solid lines are Lowess curves as visual aids. Higher scores represent more positive valence.}
\label{fig:valence_regression}
\end{figure}

Figure \ref{fig:valence_regression} shows the directly-induced valence score for each set of group labels from the ridge regression fit to the NRC VAD valence lexicon. Because we trained a different Word2Vec model for each year, we trained a different ridge regression model for each year. Over all thirty years, the Pearson correlation between predicted valence and actual valence on the test set ranged from 0.617 to 0.675, and $R^2$ values ranged from 0.423 to 0.451.The predicted scores show similar trends to the average neighbor valence. \textit{Homosexual} has the most negative valence for every year, followed by \textit{gay} and the aggregate over all LGBTQ terms, followed by \textit{American} with the most positive valence. \textit{American} is significantly more positive than all LGBTQ labels over all years (Wilcoxon signed-rank test; $p < 0.0001$), and \textit{gay} is significantly more positive than \textit{homosexual} for every year (Wilcoxon signed-rank test; $p < 0.0001$). Despite the stability in differences between these terms across experiments, the regression analysis suggests different temporal dynamics. Figure \ref{fig:valence_regression} shows that the predicted valence for \textit{gay}, \textit{homosexual}, and all LGBTQ terms all increase over time, but \textit{homosexual}'s predicted valence decreases from 2001 to 2015 ($p < 0.01$). This result is consistent with the other findings in this article in illustrating the pejoration of \textit{homosexual} in recent years.

\subsection*{Agency prediction results}

Because we use the NRC VAD dominance lexicon to quantify \textit{denial of agency} in the same way we use the valence lexicon to quantify \textit{negative evaluations of a target group}, we again directly induce scores directly from target group label representation. We train another set of ridge regression models with word embedding features for each year, but now we fit the model to the NRC VAD dominance lexicon's scores rather than the valence scores.

\begin{figure}[]
\centering
\includegraphics[width=0.75\textwidth]{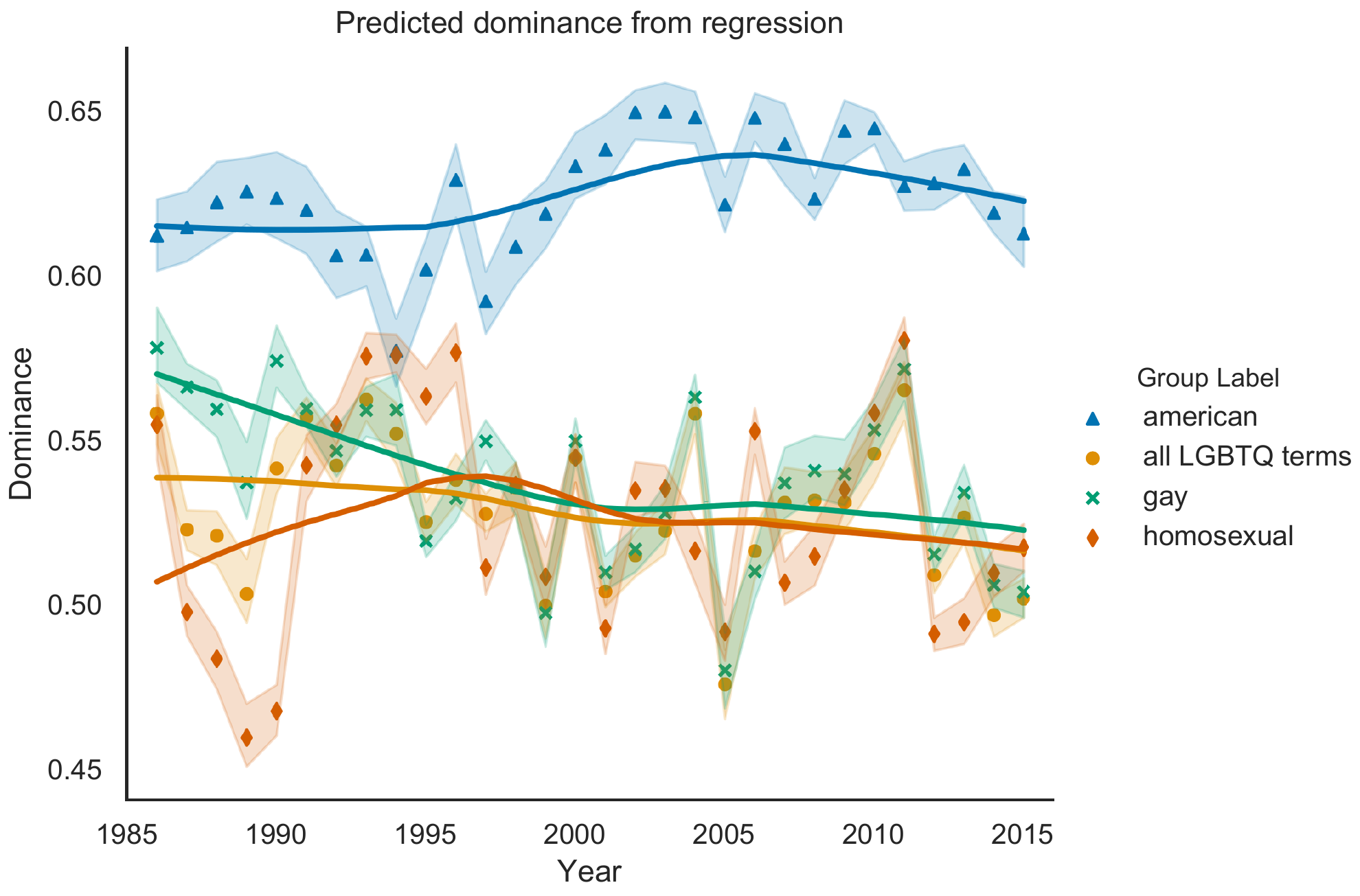}
\caption{Predicted dominance of our representations of \textit{gay}, \textit{homosexual}, \textit{all LGBTQ terms}, and \textit{American} directly induced by fitting ridge regression models to the NRC VAD Dominance Lexicon with the lexicon's words' Word2Vec representations as features for each year. Results are averaged over 10 Word2Vec models trained over each year, shaded bands represent 95\% confidence intervals and the solid lines are Lowess curves for visualization purposes. Higher scores represent greater predicted dominance. }
\label{fig:dominance_regression}
\end{figure}

Figure \ref{fig:dominance_regression} shows the predicted dominance for each group label, which is calculated by fitting ridge regression models to the NRC VAD Dominance Lexicon using the lexicon's words' Word2Vec representations as features for each year. Pearson correlations between predicted and actual dominance scores for all regression models ranged from 0.561 to 0.614 on the test set, and $R^2$ values range from 0.338 to 0.361. Consistent with the average neighbor dominance approach , \textit{American} has significantly greater dominance than any of the other LGBTQ terms (Wilcoxon signed-rank test; $p < 0.0001$). However a Wilcoxon signed-rank test over each year's means shows that there is no significant difference between the terms \textit{gay} and \textit{homosexual} ($p = 0.21$). The predicted dominance of \textit{all LGBTQ terms} and \textit{gay} significantly decrease ($p < 0.0001$), but not in the last 15 years ($p = 0.85$ for \textit{all LGBTQ terms} and $p = 0.51$ for \textit{gay}). \textit{Homosexual} does not significantly change in predicted dominance in either the full 30 years ($p = 0.96$) or in the last 15 years ($p = 0.89$). 

Why do \textit{gay} and \textit{homosexual} show such different patterns in directly-induced predicted dominance from the regression than average dominance based on their neighbors' entries in the NRC VAD lexicon? While the average dominance over the nearest neighbors showed significant differences, they were small in magnitude (often corresponding to differences of less than 0.025 points on a scale from 0 to 1). Perhaps because the word2vec features could only predict just over a third of the variance in dominance scores, they were not able to capture subtle semantic distinctions that could characterize differences in dominance scores

\end{document}